\documentclass[journal]{IEEEtran}
\usepackage{times}
\usepackage{epsfig}
\usepackage{graphicx}
\usepackage{amsmath}
\usepackage{amssymb}

\usepackage{times}
\usepackage{graphicx}
\usepackage{verbatim}
\usepackage{multirow}
\usepackage{epsfig}
\usepackage{graphicx}
\usepackage{amsmath}
\usepackage{amssymb}
\usepackage{subfigure}
\usepackage{mathrsfs}
\usepackage{color}
\usepackage{xcolor}
\usepackage{float}
\usepackage{caption}
\usepackage{pifont}
\usepackage{mathrsfs}
\usepackage{bm}
\usepackage{epstopdf}
\usepackage{algorithmic}
\usepackage{algorithm}

\newcommand{\RNum}[1]{\uppercase\expandafter{\romannumeral#1\relax}}
\newcommand{\mycaption}[1]{\caption*{\footnotesize #1}}

\newcommand{\etal}{\emph{et al. }}

\newcommand{\ie}{\emph{i.e. }}

\usepackage[colorlinks,
            linkcolor=red,
            anchorcolor=blue,
            citecolor=green
            ]{hyperref}
\ifCLASSINFOpdf
\else
\fi
\begin{document}
%
\title{Generating Superpixels for High-resolution Images \\ with Decoupled Patch Calibration}
%
%
%

\author{
Yaxiong Wang,
        Yunchao Wei, Xueming Qian,~\IEEEmembership{Member~IEEE, } Li Zhu, and Yi Yang
\IEEEcompsocitemizethanks{\IEEEcompsocthanksitem Y. Wang is with the School of Software Engineering, Xi'an Jiaotong University, Xi'an, 710049, China. He is now a visiting Ph.D student at ReLER Lab, University of Technology Sydney  \protect
 (e-mail: wangyx15@stu.xjtu.edu.cn).
\IEEEcompsocthanksitem Y. Wei is with School of Computer and Information Technology, Beijing Jiaotong University, Beijing, 100000, China
 (e-mail: wychao1987@gmail.com).\protect
\IEEEcompsocthanksitem X. Qian is with the Key Laboratory for Intelligent Networks and Network Security, Ministry of Education, Xi’an Jiaotong University, Xi’an 710049, China, also with the SMILES Laboratory, Xi’an Jiaotong University, Xi’an 710049,China, and also with Zhibian Technology Co. Ltd., Taizhou 317000, China (e-mail: qianxm@mail.xjtu.edu.cn).\protect
\IEEEcompsocthanksitem L. Zhu is with the School of Software, Xi’an Jiaotong University, Xi’an 710049, China (e-mail: zhuli@xjtu.edu.cn).\protect
\IEEEcompsocthanksitem Yi Yang is with School of Computer Science and Technology, Zhejiang University, Hangzhou, 310000, China (e-mail: yangyics@zju.edu.cn).\protect}
\thanks{Manuscript received ...; revised ....}}

\maketitle


\begin{abstract}
Superpixel segmentation has recently seen important progress benefiting from the advances in differentiable deep learning. However, the very high-resolution superpixel segmentation still remains challenging due to the expensive memory and computation cost, making the current advanced superpixel networks fail to process. In this paper, we devise Patch Calibration Networks (PCNet), aiming to efficiently and accurately implement high-resolution superpixel segmentation. PCNet follows 
the principle of producing high-resolution output from low-resolution input for saving GPU memory and relieving computation cost. To recall the fine details destroyed by the down-sampling operation, we propose a novel Decoupled Patch Calibration (DPC) branch for collaboratively augment the main superpixel generation branch. In particular, DPC takes a local patch from the high-resolution images and dynamically generates a binary mask to impose the network to focus on region boundaries. By sharing the parameters of DPC and main branches, the fine-detailed knowledge learned from high-resolution patches will be transferred to help calibrate the destroyed information. To the best of our knowledge, we make the first attempt to consider the deep-learning-based superpixel generation for high-resolution cases. To facilitate this research, we build evaluation benchmarks from two public datasets and one new constructed one, covering a wide range of diversities from fine-grained human parts to cityscapes. Extensive experiments demonstrate that our PCNet can not only perform favorably against the state-of-the-arts in the quantitative results but also improve the resolution upper bound from 3K to 5K on 1080Ti GPUs.

\end{abstract}


\section{Introduction}
Superpixel segmentation targets assigning the pixel with similar color or other low-level properties to the same groups, which could be viewed as a clustering procedure on the image, the formed pixel clusters are known as superpixels. Benefiting from the tremendous progress of deep convolution neural networks, many approaches have been proposed to harness deep models to facilitate superpixel segmentation and achieved promising results~\cite{SEAL,SCN,SSN}. The common practice first evenly splits the image into grids and utilize the convolution network to predict a 9-dimension vector for each pixel, which indicates the probabilities of the pixel assigned to its 9 surrounding grids~\cite{SCN,SSN}.  Nevertheless, existing methods often fail to process ultra high-resolution images due to memory limitation of GPUs.
\begin{figure}[t]
\begin{center}
\subfigure{
    \begin{minipage}[t]{0.96\linewidth}
        \centering
        \includegraphics[height=2.5in,width=3.25in]{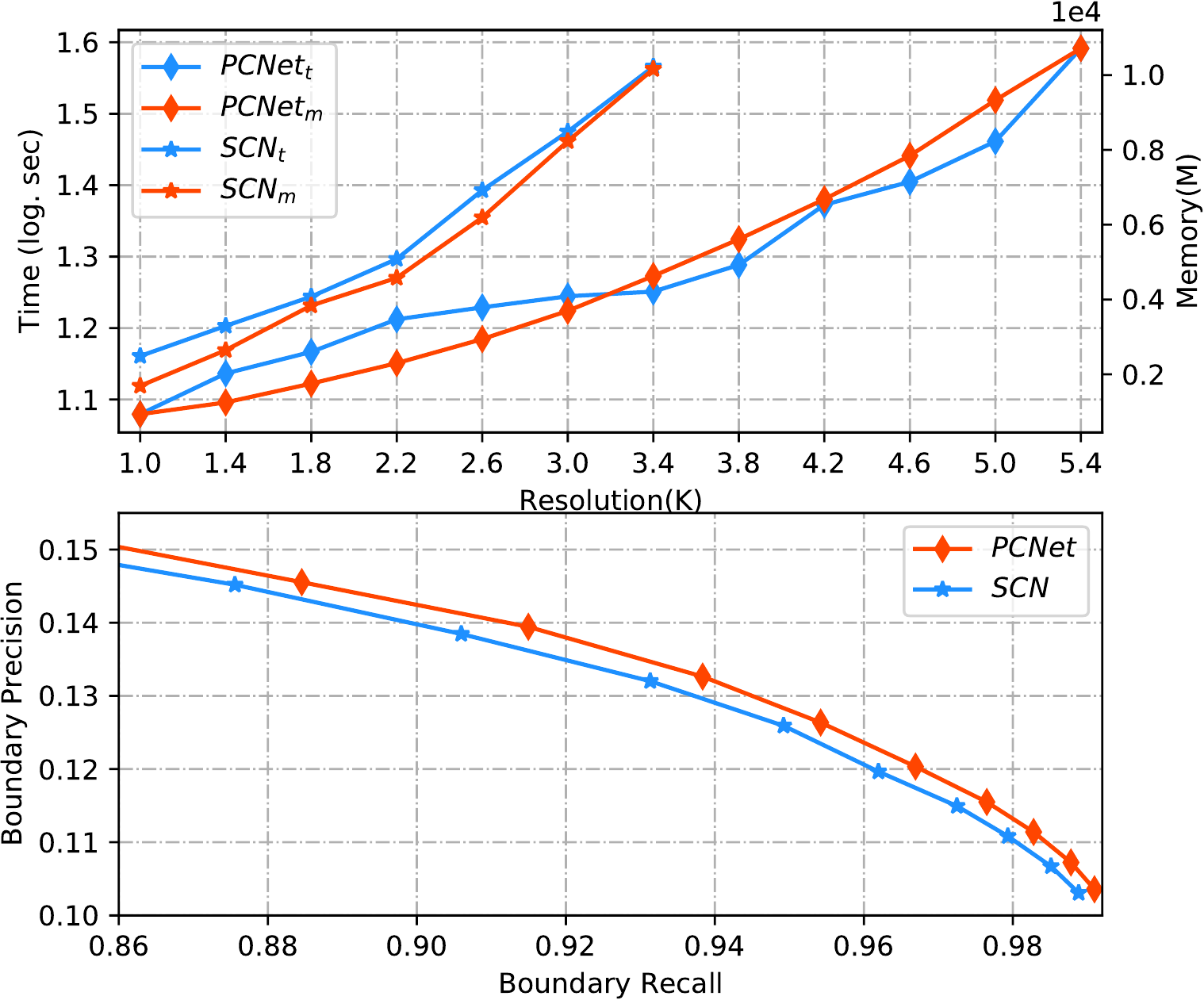}\\
    \end{minipage}
    }
\vspace{-0.2cm}
\caption{Time \& memory, and performance comparison between our PCNet and the state-of-the-art SCN~\cite{SCN}, the footnote `t' and  `m' indicates the time and memory cost, respectively. Our method could efficiently acquire superpixels of high-resolution images (5K) while keep competitive boundary precision. Results are evaluated using a single NVIDIA 1080Ti GPU.}
\vspace{-0.2cm}
\label{fig1}
\end{center}
\end{figure}

\begin{figure*}
\begin{center}
\subfigure{
    \begin{minipage}[t]{0.98\linewidth}
        \centering
        \includegraphics[height=1.3in,width=6.8in]{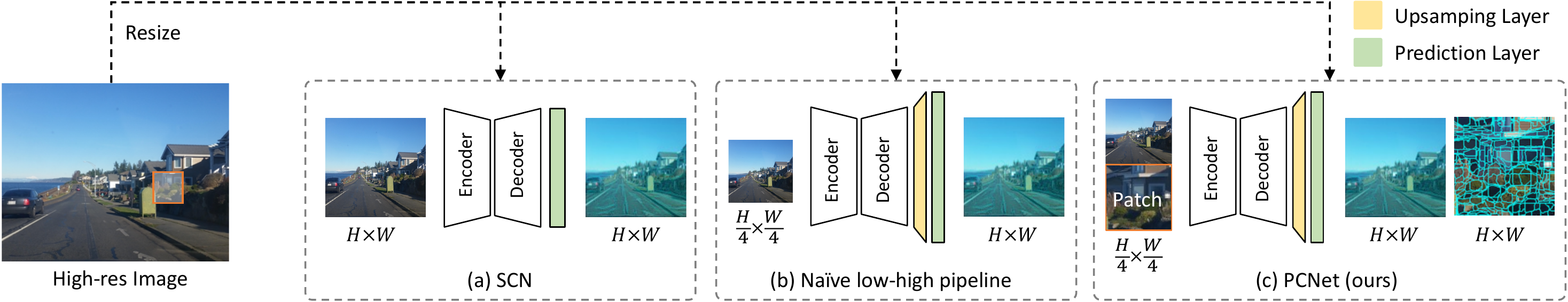}\\
    \end{minipage}
    }
    \vspace{-0.1cm}
\caption{The comparison of different training architectures, SCN~\cite{SCN} outputs the prediction whose resolution is the same as the input. The na\"\i ve solution directly predicting high-resolution from the low-resolution image would result in a poor performance. By introducing calibration branch, our PCNet could process higher resolution images and achieve a satisfactory performance at the same time.}
\vspace{-0.2cm}
\label{argue}
\end{center}
\end{figure*}
Ultra high-resolution problem is a longstanding challenge in computer vision, especially for the pixel-wise tasks like segmentation~\cite{cascadePSP,deepStrip,HRMatting,HRNet}, optical flow~\cite{optical_flow1,optical_flow2,optical_flow3,optical_flow4}, depth estimation~\cite{depth_estimation2,depth_estimation3,depth_estimation4,depth_estimation5}~\etal.  For superpixel segmentation, although some deep learning based methods have been proposed, the ultra high-resolution scenario has not been well explored. For instance, the state-of-the-art method, SCN~\cite{SCN}, could generate superpixels for low-resolution images, however, when encountering images with higher resolution, its inference speed will be much slower. What's worse, SCN could not work when the resolution of the fed image is over 3.5K based on NVIDIA 1080Ti GPU, as shown in Fig.~\ref{fig1}.
To enlarge the tolerate resolution of SCN, a straightforward solution is to predict the high-resolution output from a low-resolution image. Specifically, by introducing additional upsampling layers in the decoder stage, we could enforce the network to learn the  $H\times W$  association map from a lower-resolution $H/4\times W/4$ image, as shown in Fig.~\ref{argue} (b).  With such a design principle, the network could successfully process larger size images. 

However, too much sacrificed information in the very low-resolution input would obstruct the network to perceive the textural contexts, especially the fine boundary details.
As a result, the performance would remarkably deteriorate. For perceiving more boundary details, we propose to calibrate the boundary pixels assignment by directly cropping a patch from the original high-resolution images as additional input, leading to the \textbf{P}atch \textbf{C}alibration Network (PCNet).  The core idea of our PCNet is illustrated in Fig.~\ref{argue} (c), we propose to crop a local patch from the source high-resolution image such that all details could be reserved. We then feed it into the Decoupled Patch Calibration (DPC) branch to perform a calibration for the boundary details of the coarse global prediction. By sharing weights with the main branch, the learned knowledge from DPC could be well transferred to make the main superpixel branch accurately perceive more boundary pixels.  

Different from the global input that attempts to perceive the overall boundary layout by classifying the pixels to their object categories, our cropped patches only target helping the network accurately assign the pixels around the boundaries while paying no attention to the included objects. To impose this point, a dynamic guiding training mechanism is designed. Instead of greedily identifying the categories of all pixels in the patch, we design a dynamic mask to encourage the network to only focus on the main boundaries of current patch,  which is achieved by degrading the multi-class semantic label to a dynamic binary mask as guidance.
With our dynamic guiding training, in each iteration, the network only needs to discriminate the foreground and the background to highlight the main boundary while spare no efforts to identify the multiple categories for all pixels. Such a strategy could not only ease the network optimization but enable the network to accurately identify more boundary pixels. 

 

From Fig.~\ref{fig1}, our PCNet could successfully process the \textbf{5K} resolution images on a single NVIDIA 1080Ti GPU. Comparing to the state-of-the-art SCN, PCNet could surpass it by a small margin even though receiving a 4-time smaller resolution input. Quantitative and qualitative results on Mapillary Vistas~\cite{vistas}, BIG~\cite{cascadePSP} and our created Face-Human datasets demonstrate that the proposed method could effectively process high-resolution $5K$ images and achieve more outstanding performance.  In summary, we make the following contributions in this work:

\begin{figure*}
    \centering
    \includegraphics[height=2.4in, width=6.5in]{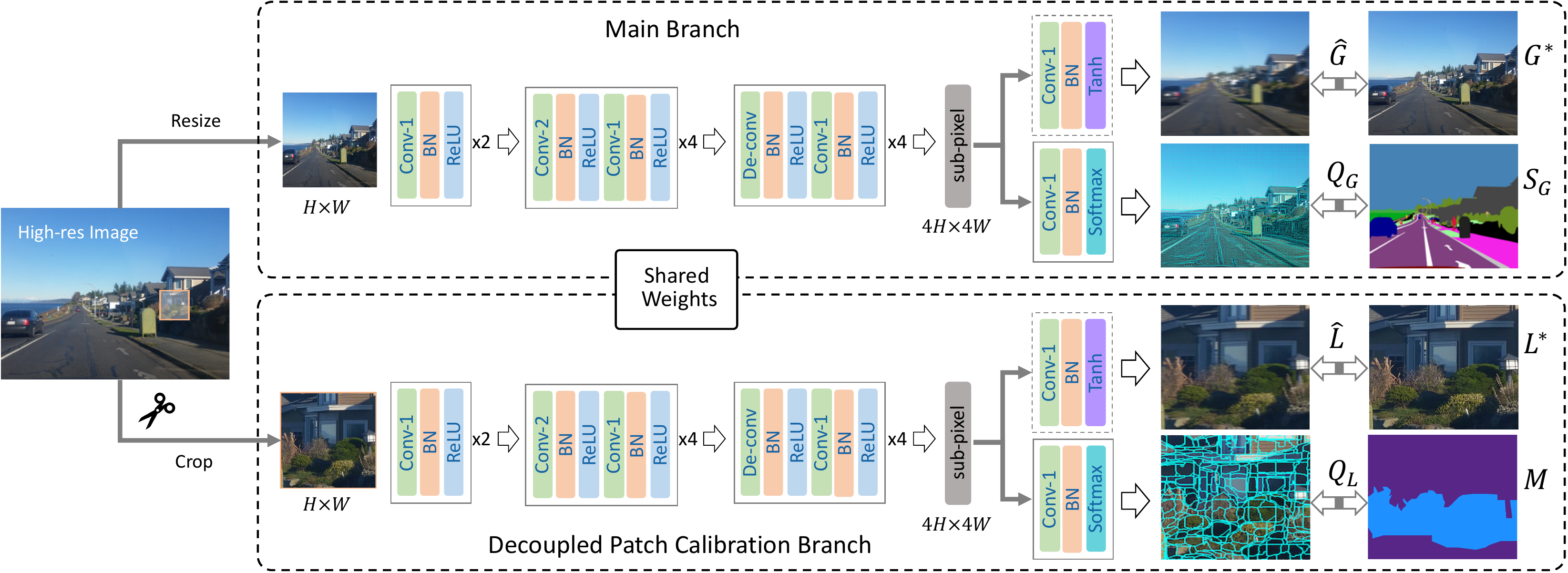}
    \caption{Illustration of proposed PCNet. In each iteration, the low-resolution global and local inputs are fed forward the network and predict the 4-time resolution associations, which are supervised by the ground-truth semantic labels and our dynamic guiding mask, respectively.  The super-resolution branch serves as an auxiliary module to help recover more texture details during training and is discarded when inference. And `conv-\#' indicates the convolution with stride \#, the `sub-pixel' stands for the sub-pixel convolution operation with scale 4~\cite{pix_shuffle}. The `SR' and `SP' mean the super-resolution and superpixel heads, respectively. }
   \vspace{0.3cm}
    \label{workflow}
\end{figure*}

\begin{itemize}
    \item We contribute the first framework for high-resolution superpixel segmentation. The proposed PCNet could effectively process 5K images and achieves satisfactory performance on three benchmarks.
    
    \item A novel patch calibration training paradigm is proposed, with this architecture, the global image and the local patch could compensate each other and work together to train a more robust model. 
    
    \item We design a dynamic guidance training method, a binary mask is dynamically generated and guide the network to focus on the boundaries of semantic regions, which could efficiently benefit the boundary identification. 
    
\end{itemize}

In the following, we would first introduce the existing works related to this paper in section~\ref{related} and elaborate the details of our PCNet in section~\ref{method}. Section~\ref{experiment} would present the experimental results and make comparison with the state-of-the-art methods. Finally, the conclusions would be given in section~\ref{conclusion}. 

\section{Related Work}
\label{related}
Although the high-resolution superpixel segmentation has not been well studied, the superpixel segmentation for general images has a long line of research and has made important progress in recent years. Besides, the high-resolution segmentation is a neighbor tasks for the high-resolution superpixel segmentation, since the superpixel could be viewed as over-segmentation of image. In the following two subsections, we will subsequently present the existing works of superpixel segmentation and high-resolution segmentation, respectively.

\subsection{Superpixel Segmentation}
Superpixel segmentation can be viewed as a clustering procedure on the image, the key of this problem is to estimate the assignment of each pixel to its potential clustering centers~\cite{ERS,SLIC,MSLIC,SEAL,SCN,SNIC,add_sp1,add_sp2,sp_app1,sp_app2,sp_app3,sp_app4}. Traditional methods model the assignment using the clustering theory~\cite{LSC,SLIC,MSLIC,SNIC} or graph technique~\cite{FH,ERS}. In general, the clustering-based methods usually use the clustering strategy to compute the connectivity between the anchor pixel and its neighbors, which is straightforward for superpixel segmentation. The well-known superpixel method SLIC~\cite{SLIC} adapt the $k$-means for superpixel segmentation and is a classic algorithm in superpixel community. Liu \etal~\cite{MSLIC} extend the SLIC to compute content-sensitive superpixels, the designed model could generate the small superpixels in content-dense regions and large superpixels in content-sparse regions. Li \etal~\cite{LSC} explicitly utilize the connection between the optimization objectives of weighted $k$-means and normalized cuts by introducing a elaborately designed high deimensional space.  While the graph-based approaches formulate the superpixel segmentation as a graph-partitioning problem, and the superpixel segmentation is performed by estimating the connectivity strength between the pixels.  FELZENSZWALB \etal~\cite{FH} utilize a graph-based representation of the image and define a predicate for measuring the evidence for boundary between two regions, the authors design an efficient superpixel segmentation algorithm named \emph{FH} based on the predicate. In~\cite{ERS}, the Liu \etal propose a novel objective function for superpixel segmentation, and the segmentation is then given by the graph topology that maximizes the objective function under the matroid constraint. Inspired by the success of deep neural networks on many vision tasks, researchers recently attempt to harness deep convolutional networks to facilitate superpixel segmentation. Tu \etal~\cite{SEAL} propose to  improve the superpixel segmentation by the learned pixel-wise deep features from fully connected networks. Jampani \etal~\cite{SSN} propose a soft clustering mechanism and incorporate it with deep architecture, they develop the first end-to-end deep solution for superpixel task. In~\cite{SCN}, the authors further simplify the framework in~\cite{SSN} and contribute a faster superpixel segmentation algorithm.

\subsection{High-resolution Segmentation}
The high-resolution superpixel segmentation is rarely explored, while its neighbor task, segmentation for high-resolution images has been studied by many works~\cite{cascadePSP,HRseg1,HRseg2}. Cheng \etal~\cite{cascadePSP} fuse the multi-scale information and iteratively refine the given coarse segmentation map to produce a high-resolution prediction. Chen \etal~\cite{HRseg1} utilize two branches to capture the global and the local context, which are fused to predict the final segmentation, the proposed framework is memory-efficient. In~\cite{HRseg3}, Sarker \etal propose a prior and sub-strips based mechanism to address the very high-resolution image segmentation.  Instead of giving high-resolution prediction from low resolution, Wang \etal design a parallel and hierarchic architecture, named HRNet~\cite{HRNet}, to maintain high-resolution information through the whole process. Lin \etal~\cite{HRseg4} maintain a long-range residual connection to exploits all the information available along the down-sampling process, which could enable the network to capture the high-level semantic features and the fine-grained textual features simultaneously. In~\cite{HRseg5}, Zhao \etal utilize a cascade paradigm to incorporate multi-resolution branches for high-resolution image segmentation, the authors introduce a cascade features fusion unit to perform high-quality segmentation. Zheng \etal~\cite{HRSeg} employ a global-local joint learning strategy to perform efficient segmentation for ultra high-resolution remote sensing images. In~\cite{HRSeg_add1}, Zhou \etal focus on the medical images and propose a novel high-resolution multi-scale encoder-decoder network, the authors introduce multi-scale dense connections to exploit comprehensive semantic information. To address the high-resolution segmentation, they integrated a high-resolution pathways to collect the high-resolution semantic information for boundary localization. Liao \etal~\cite{HRSeg_add2} further explore the segmentation of 3D high-resolution magnetic resonance angiography (MRA) images of the human cerebral vasculature and develop two minimal path approaches. Li \etal propose an edge-embedded marker-based watershed algorithm for high spatial resolution remote sensing image segmentation~\cite{HRSeg_add3}, the authors utilize the confidence embedded method to detect the edge information and propose a two-stage model, where the first stage extract maker image while the second stage integrates the edge information into the labeling process with the edge pixels assigned the lowest priority and lastly labeled,  such that the edge pixels become the candidates of the boundary pixels and more precise objects boundaries can be acquired. 

\section{Methodology}
\label{method}

Fig.~\ref{workflow} shows the schematic illustration of our proposed PCNet. In the training stage, two types of inputs, \ie, the global image $G\in \mathcal{R}^{3\times H\times W}$ and the cropped local patch $L\in \mathcal{R}^{3\times H\times W}$, are fed into the main branch and the decoupled patch calibration branch, respectively.
The inputs are down-sampled 4 times in the contracting path and upsampled 6 times in the expansive path. Consequently, both branches output association maps with shape $9 \times 4H\times 4W$, which are supervised by the semantic label and our designed dynamic guiding mask. A super-resolution branch serves as the auxiliary module to help the network restore more details from the low-resolution inputs, and is discarded during inference. We share weights between two branches to propagate the learned knowledge.  

\subsection{Patch Calibration Arch.}
As shown in Fig.~\ref{workflow}, for memory and computation efficiency, our PCNet predicts the high-resolution $4H\times 4W$ association maps with low resolution $H\times W$ inputs using an encoder-decoder paradigm.  Instead of using the symmetric layers in encoder and decoder, we additionally introduce a sub-pixel layer~\cite{pix_shuffle} with scale factor 4 before the prediction layer to help produce an association map $Q$ with shape $9\times 4H\times 4W$. During training, the main branch is responsible to distill the global boundary layout from $G\in \mathcal{R}^{3\times H\times W}$, which is obtained by simply resizing the original high-resolution image.  While the decoupled patch calibration (DPC) branch focuses on calibrating the global results by capturing finer boundaries from the cropped local patch $L$.  By sharing the learned weights of DPC branch, the main branch could simultaneously perceive the panoramic boundary layout and the fine boundary details, efficiently preventing the performance from deteriorating.

The outputs of the two branches are supervised in a similar manner. Give the association prediction $Q_G$ and the corresponding label $S_G$ of the global input, the superpixel training is performed by first computing  the center of any superpixel $s$ from the surrounding pixels:
\begin{equation}
\label{step1}
    h(s) = \frac{\sum_{p:s\in N_p}{S_G(p)\cdot Q_G(p,s)}}{\sum_{p:s\in N_p}Q_G(p,s)},
\end{equation}
and then reconstruct the property of any pixel according to the superpixel neighbors:
\begin{equation}
\label{step2}
    S_G^{'}(p) = \sum_{s\in N_p} h(s)\cdot Q_G(p,s),
\end{equation}
where the $N_p$ is the set of adjacent superpixels of $p$, $Q_G(p,s)$ indicates the probability of pixel $p$  assigned to superpixel $s$. Thus, the network is optimized towards minimizing the distance between the ground-truth label and the reconstructed one. Following Yang~\etal~\cite{SCN}, the 2-dimension spatial coordinate $p$ is also considered, thus, the full superpixel loss reads:
\begin{equation}
\label{sp_loss}
    \mathcal{\bm{L}_{SP}}(Q_G,S_G) = \sum_{p} CE(S_G^{'}(p), S_G(p)) + ||p^{'}-p||_2^{2}
\end{equation}
where $CE(\cdot,\cdot)$ stands for the cross entropy loss, and $p^{'}$ is the reconstructed vector from $p$ according to Eq.~\ref{step1}-~\ref{step2}. 

To restore more details from the low-resolution input $G$, a super-resolution branch is further introduced to give a reconstruction $\hat{G}$ of the high-resolution version $G^*$, which is optimized by a masked $L_1$ reconstruction loss, the details would be elaborated in subsection~\ref{loss_functions}. The full loss for global branch is: $\mathcal{L}_G=\mathcal{L}_{SP}(Q_G, S_G) + \mathcal{L}_{SR}(\hat{G},G^*)$

\begin{figure}
    \centering
    \includegraphics[height=1.7in, width=2.4in]{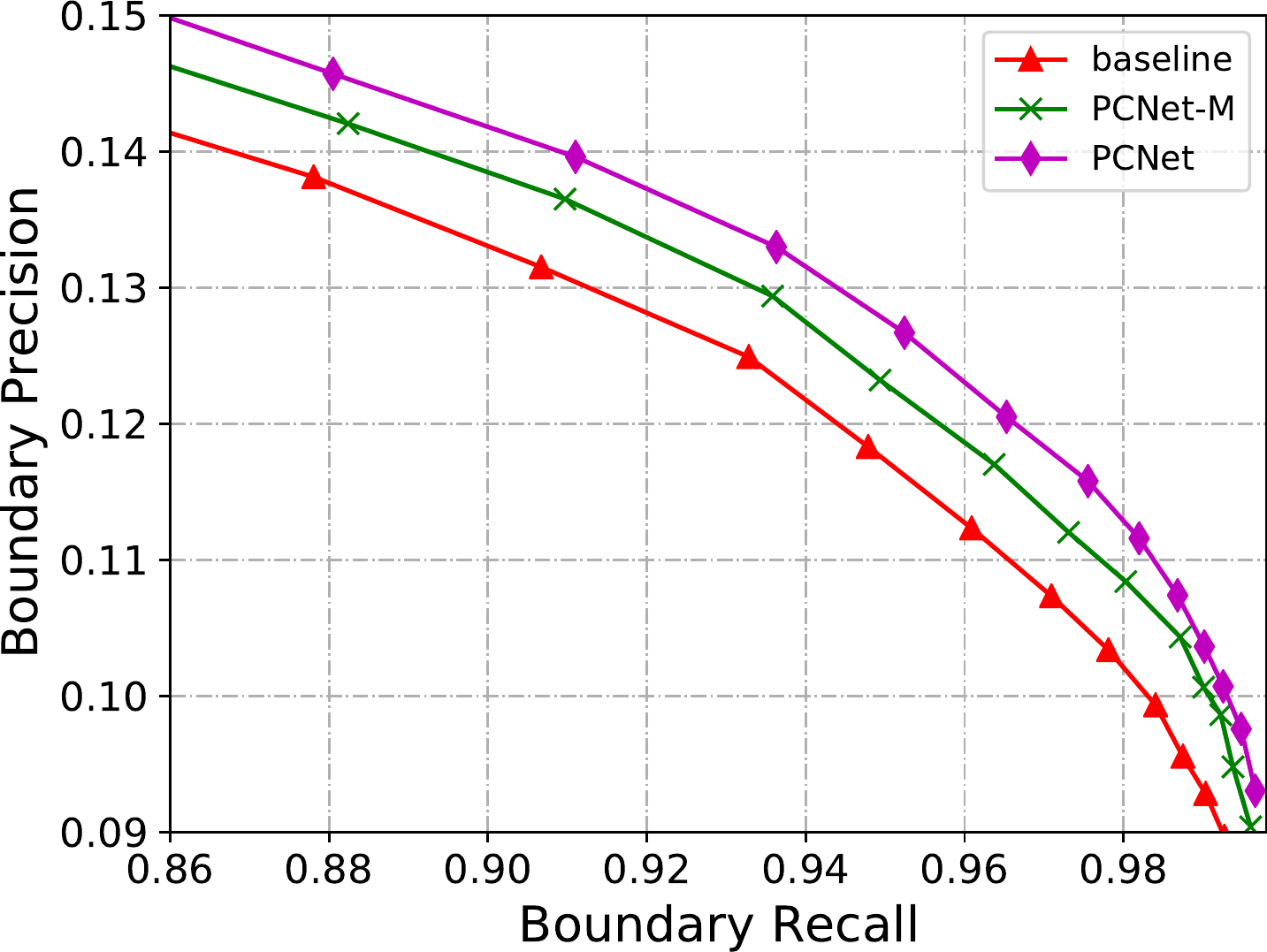}
    \vspace{-0.2cm}
    \caption{Performance comparison of PCNet with the multi-class label and the dynamic mask.}
    \vspace{0.2cm}
    \label{binary_mask_validation}
\end{figure}

\subsection{Decoupled Patch Calibration Branch}

Our PCNet attempts to produce high-resolution outputs from low-resolution inputs,  such a paradigm allows us to train a network that could tolerate higher resolution images. However, the lost textural details and the blur boundary contexts in the down-sampled input make the network difficult to well identify the boundary pixels. To remedy this weakness, we design our decoupled patch calibration (DPC) branch. Particularly, the local patch $L\in\mathcal{R}^{3\times H\times W}$ is first cropped from the original high-resolution data anchored on a boundary pixel, and forward through the DPC branch to capture finer boundary details. Finally, we endue the main branch with the ability of fine boundary perceiving by sharing the learned weights of DPC.  

Given the output $Q_L$ of DPC, the ground-truth semantic label $S_L$ is a straightforward choice to serve as the supervision, which attempts to benefit the superpixel segmentation by classifying the pixel to its object category. This strategy could success on global images but did not work well on the local input, since the cropped patch usually only covers a part of the objects in our practice, as shown in Fig.~\ref{workflow}. The lack of global context makes the multi-class classification procedure much difficult. Actually, unlike the semantic segmentation that needs to capture the global context to identify complete objects, the superpixel segmentation mainly concerns whether the boundaries are accurately identified~\cite{SCN,SSN,SP_evaluation}. In other terms, the superpixel network only needs to discriminate the adjacent regions for boundary perceiving, and it's unnecessary to identify all object categories of the pixels. 
Motivated by this consideration, we propose our dynamic guiding mask to supervise the local association prediction. Formally, we first sample the semantic label corresponding to the local patch and find the class $c$ with the longest boundaries. Then, the dynamic guiding mask is defined as follows:
\begin{equation}\label{dym_mask}
M(p)=\left\{ \begin{aligned}
    1 &  &\text{if  } S_L(p)=c,\\
    0 & &\text{otherwise.\quad}
\end{aligned}
\right.
\end{equation}

\setlength{\textfloatsep}{7pt}
\renewcommand{\algorithmicrequire}{\textbf{Input:}}
\renewcommand{\algorithmicensure}{\textbf{Output:}}
\begin{algorithm}[t]
	\caption{\small Training pseudocode of PCNet (PyTorch-like).}
	\begin{algorithmic}[1]
		\REQUIRE The training set, network PCNet with parameters $\Theta$, optimizer($\Theta$), hyperparameters  $\alpha, \beta$;
		\ENSURE optimized network parameters $\Theta^{*}$   \\
		\WHILE{not converged}
		\FOR {$I, S \in \text{training set}$}
		\STATE  \textcolor{gray}{\# prepare inputs, $b$ is a boundary pixel}
		\STATE  $G^*$, $S_G$, $L^{*}$ = $\text{Resize}(I)$, $\text{Resize}(S)$, $\text{Crop}(I, b)$;
		\STATE  Generate dynamic mask $M$ from Eq.~\ref{dym_mask};
		\STATE  $G$, $L$ = $\text{Scale}(G^*, \frac{1}{4})$, $\text{Scale}(L^*, \frac{1}{4})$;
		\STATE  \textcolor{gray}{\# forward}
		\STATE  $\hat{G},Q_G, E$ = PCNet($G$); \textcolor{gray}{\# for global input}
		\STATE $\hat{L},Q_L$ = PCNet($L$); \textcolor{gray}{\# for local input}
		\STATE  \textcolor{gray}{\# loss compuatation}
		\STATE  Compute $\mathcal{L}_{SP}(Q_G, S_G)$, $\mathcal{L}_{SP}(Q_L, M)$ from Eq.~\ref{sp_loss};
		\STATE  Compute $\mathcal{L}_{SR}(\hat{G},G^*)$, $\mathcal{L}_{SR}(\hat{L}, L^*)$ from Eq.~\ref{recon_loss};
		\STATE  Compute $L_\mathcal{D}(E)$ from Eq.~\ref{ld_loss}-~\ref{full_loss}; 
		\STATE  \textcolor{gray}{\# accumulate the above losses as full loss}
		\STATE  $\mathcal{L} = \mathcal{L}_G + \alpha\mathcal{L}_L +\beta\mathcal{L_D}$
		\STATE \textcolor{gray}{\# backward and update parameters $\Theta$}
		\STATE  $\mathcal{L}.$backward()
		\STATE  optimizer.step()
		\ENDFOR \\
		\ENDWHILE
	\end{algorithmic}
\end{algorithm}

With our dynamic guiding mask, the multi-class object recognition is degraded to the salient region detection,  such that the network only needs to discriminate the class $c$ and other classes, which eases the network optimization. If the foreground category $c$ could be well perceived,  most boundaries of current patch could be identified due to the choice of class $c$. For the ignored boundaries, they could be highlighted in other iterations, since the local patches are randomly cropped and the guiding mask is dynamically generated. In our practice, our dynamic guiding mask could contribute much more performance gains comparing the mutli-class label $S_L$, as shown in Fig.~\ref{binary_mask_validation}. It is worth noting that the dynamic guiding training strategy is not suitable for the main branch, since the boundary of global input is rich enough, too many boundaries would be ignored when performing Eq.~\ref{dym_mask}. What's more important, the randomness of $G$ is much smaller, which means the longest class $c$ could be the same with high probability in different iterations, as a result, the ignored boundaries could not be well captured during training.

At training stage, the dynamic binary mask $M$ would replace $S_L$ to supervise the local output $Q_L$ in Eq~\ref{sp_loss}. The super-resolution is also employed as an auxiliary branch in DPC to help restore more textures. Thus, the full loss for DPC branch is: $\mathcal{L}_L=\mathcal{L}_{SP}(Q_L, M) + \mathcal{L}_{SR}(\hat{L},L^{*})$, where $\hat{L}\in \mathcal{R}^{3\times 4H\times 4W}$ are the local patch reconstruction and $L^{*}\in \mathcal{R}^{3\times 4H\times 4W}$ is the high-resolution version of $L$.

\subsection{Training}
\label{loss_functions}
The network is trained by a combination of superpixel loss, super-resolution loss, and our proposed local discrimination loss. The super-resolution loss is a masked $l_1$ reconstruction loss:
\begin{equation}
\label{recon_loss}
    \mathcal{\bm{L}_{SR}}(G^*,\hat{G})= \frac{1}{|B|} ||B\odot (G^{*}-\hat{G})||_1,
\end{equation}
where $B$ is the binary mask to indicate the boundary pixels. To obtain $B$, we first extract the boundaries from the ground-truth semantic label and dilate it with $16\times 16$ kernel to include more boundary contexts.

Besides the superpixel loss and the super-resolution loss, we further design a local discrimination loss to highlight the boundary pixels in hidden-feature level.  
To be specific, let $E\in \mathcal{R}^{H\times W\times D}$ be the pixel-wise embedding map produced by the sub-pixel layer. Since the ground-truth label is available during training, we could sample a small local patch $B\in \mathcal{R}^{K\times K\times D}$ surrounding a boundary pixel from $E$. For simplicity, we only sample the patches covering two different semantic regions, \ie,  $B$ is a groups of features from two categories: $\{f_1,\cdots, f_m, g_1, \cdots, g_n\}$, 
where $f, g\in \mathcal{R}^{D}, m+n=K^2$. 
Intuitively, we attempt to make the features in the same categories be closer, while the embeddings from different classes should be far away from each other. To this end, we evenly partition the features in the same categories into two groups, $\bm{f}^1, \bm{f}^2, \bm{g}^1, \bm{g}^2$, and minimize the intra-dispersion while maximize the inter-dispersion:
\begin{equation}
\label{ld_loss}
    \mathcal{\bm{L}_B} = -\frac{||\bm{\mu}_f-\bm{\mu_g}||^2_2}{\bm{S}_f^2 + \bm{S}_g^2},
\end{equation}
where $\bm{\mu}_f$ and $\bm{S}_f$ are the average representation and compactness measure for features $\{f_i\}_{i=1}^m$:
    $\bm{\mu}_{f} = \frac{1}{|\bm{f}|}\sum_{f\in \bm{f}} f,
    \bm{S}_f = \sum_{f\in \bm{f}} ||f-\bm{\mu_f}||_2^2$.
Taking all of the sampled patches $\mathcal{\bm{B}}$ into consideration, the local discrimination loss is formulated as:
\begin{equation}
\label{full_LD}
    \mathcal{\bm{L}}_{\mathcal{D}} = \frac{1}{|\mathcal{\bm{B}}|}\sum_{B\in\mathcal{\bm{B}}} {\mathcal{L}_B}.
\end{equation}
By this loss, the boundary pixels could be distinguished beforehand in hidden-feature level, which could ease the following superpixel module to identify the semantic boundaries. In our practice, equipping the LD loss on the local embedding did not contribute more performance gains, therefore, our LD loss only acts on the global embedding.

\begin{figure}[t]
\setlength{\abovecaptionskip}{-1pt} 
\begin{center}
   \subfigure[Images]{
    \begin{minipage}[t]{0.45\linewidth}
        \centering
        \includegraphics[width=1.55in]{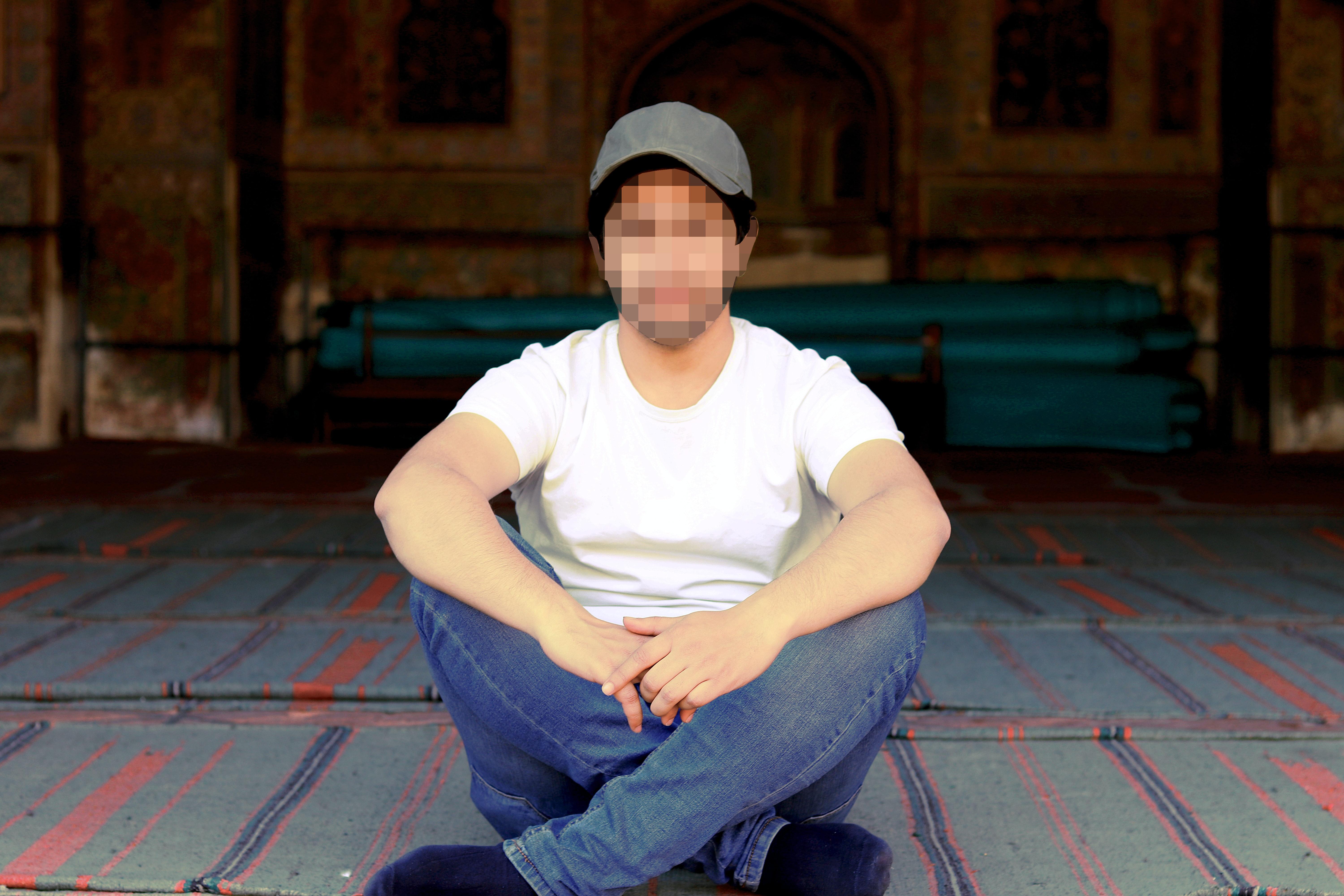}\\
        \includegraphics[width=1.55in]{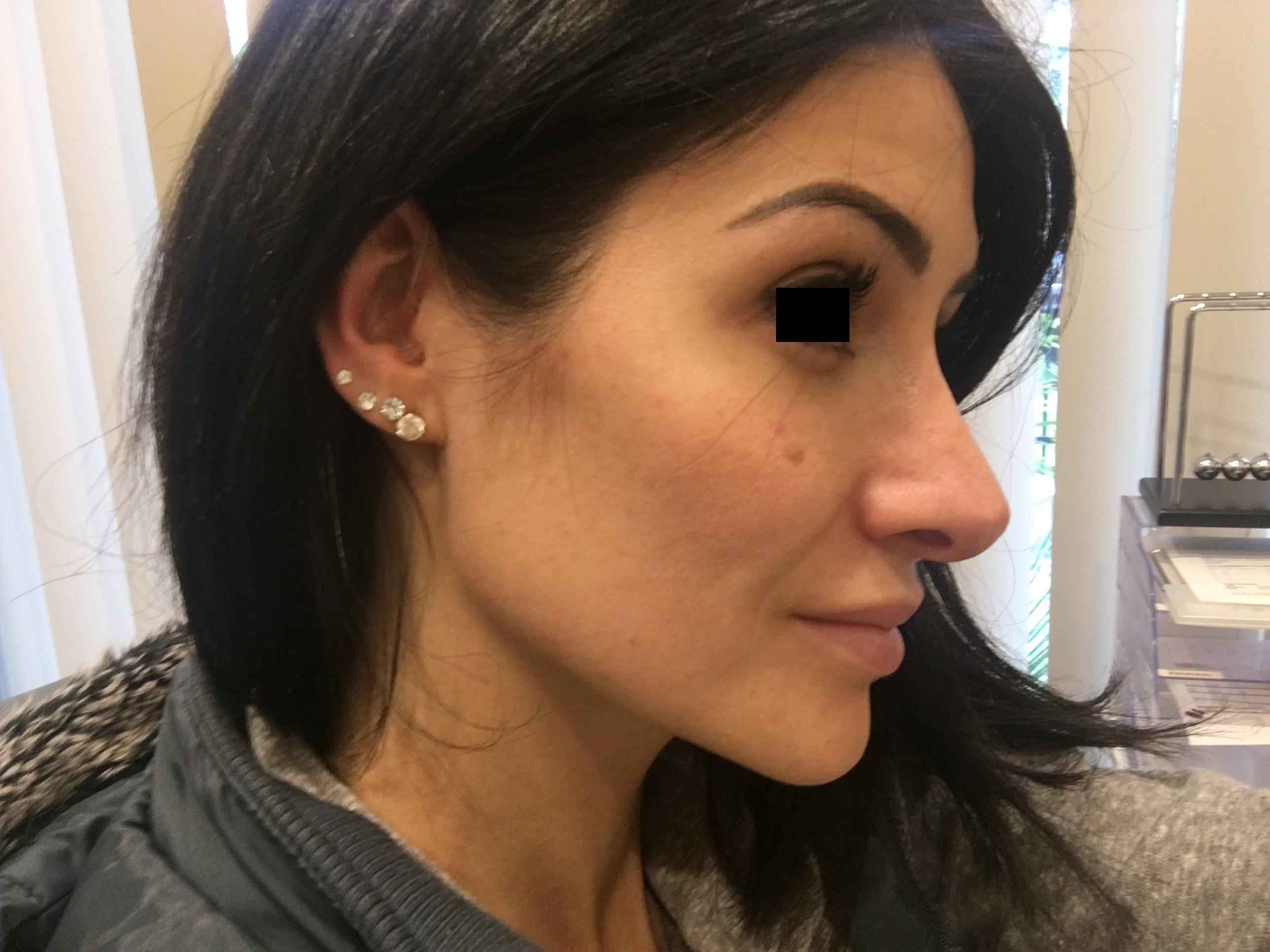}\\
        \vspace{0.1cm}
    \end{minipage}
    \label{fig:fig3_a}
    }
\subfigure[Annotations]{
    \begin{minipage}[t]{0.45\linewidth}
        \centering
        \includegraphics[width=1.55in]{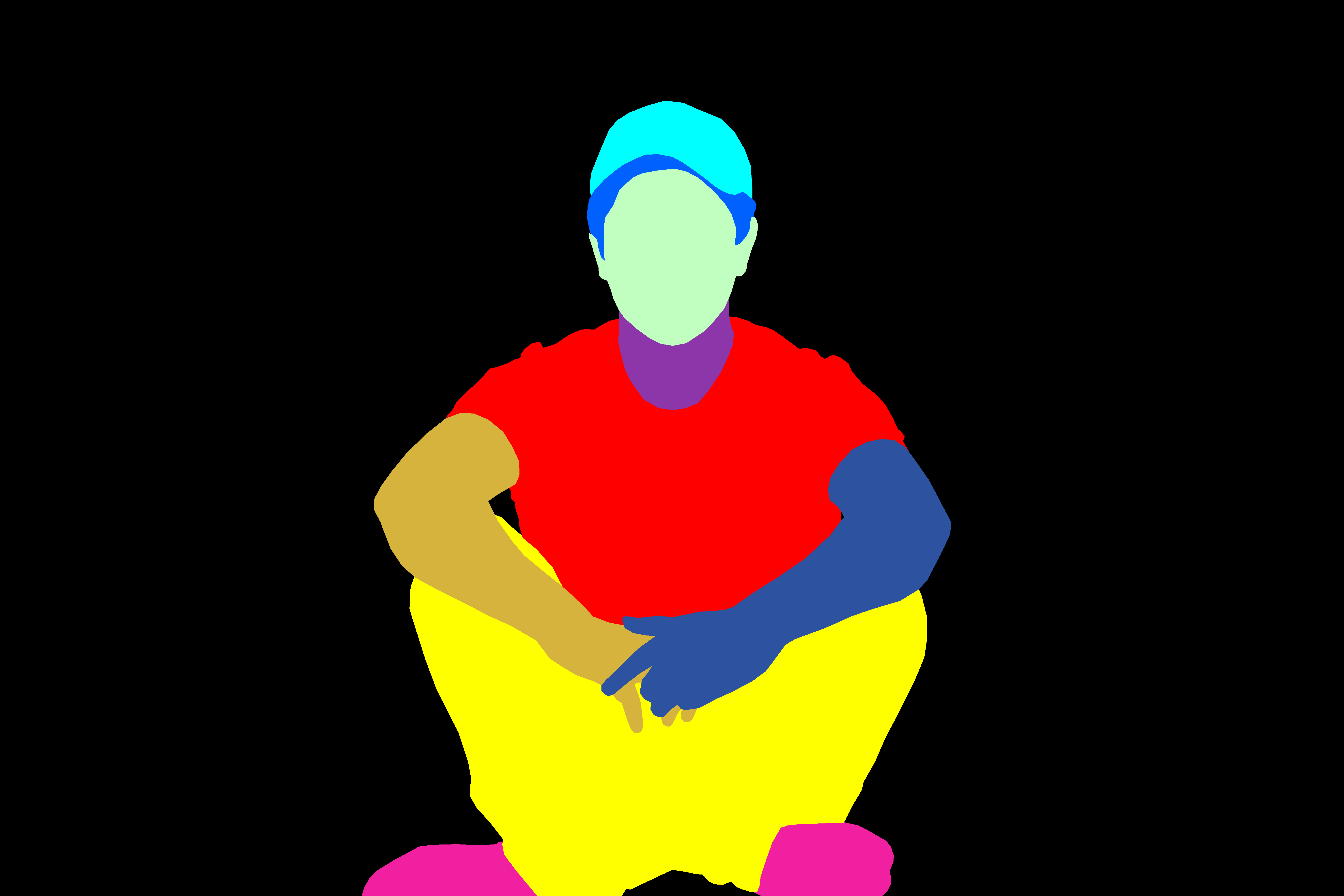}\\
        \includegraphics[width=1.55in]{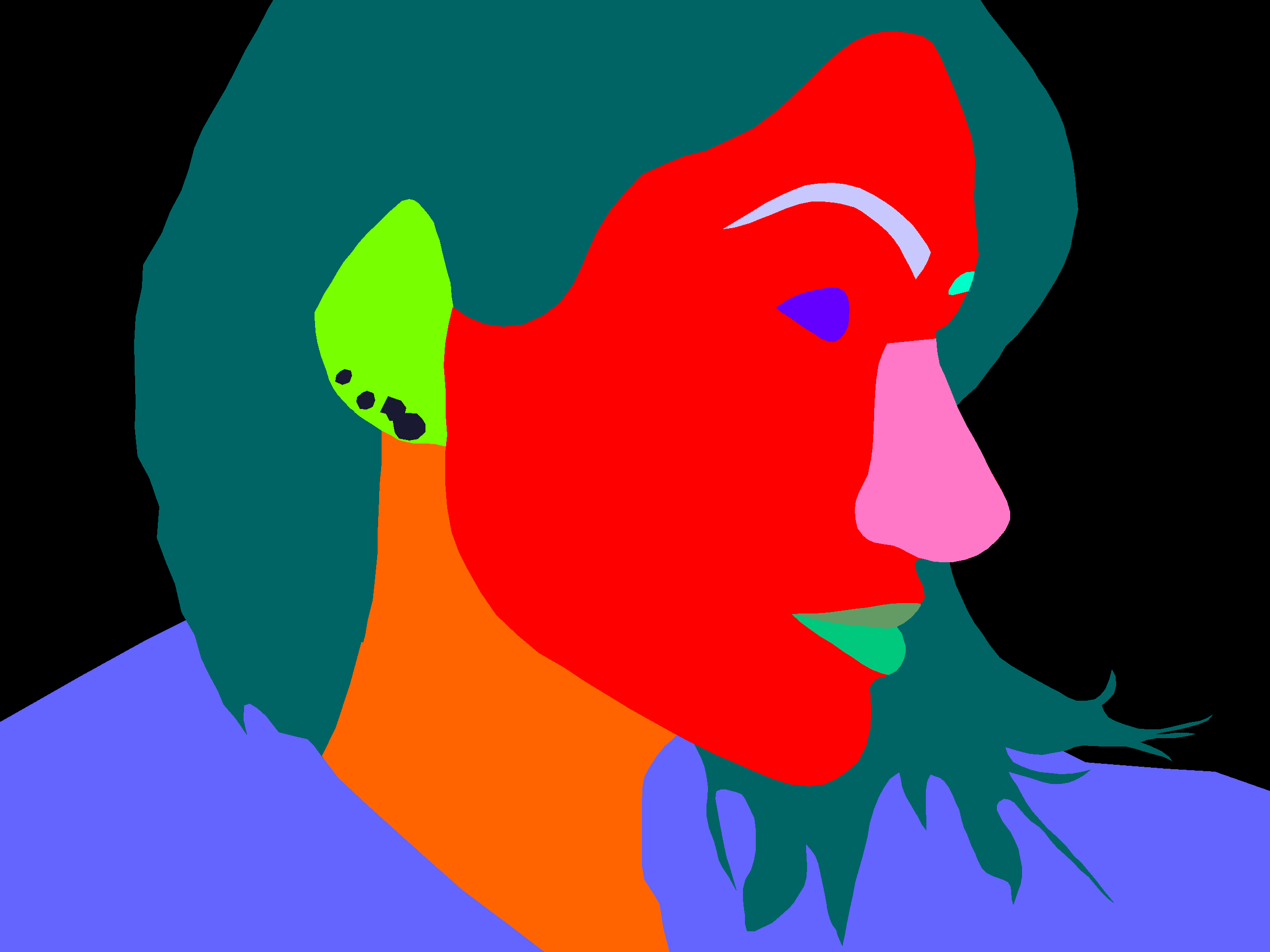}\\
          \vspace{0.1cm}
    \end{minipage}
    \label{fig:fig.3_c}
    }
\end{center}
\vspace{-0.15cm}
   \caption{Two exemplar samples from the collected Face-Human dataset, the top and the bottom rows respectively exhibit a human sample and a face sample. The key identity-regions are masked for privacy security.}
   \vspace{0.6cm}
\label{data_show}
\end{figure}

\begin{figure*}[t]
\begin{center}
\subfigure{
    \begin{minipage}[t]{0.28\linewidth}
        \centering
        \includegraphics[width=2.1in]{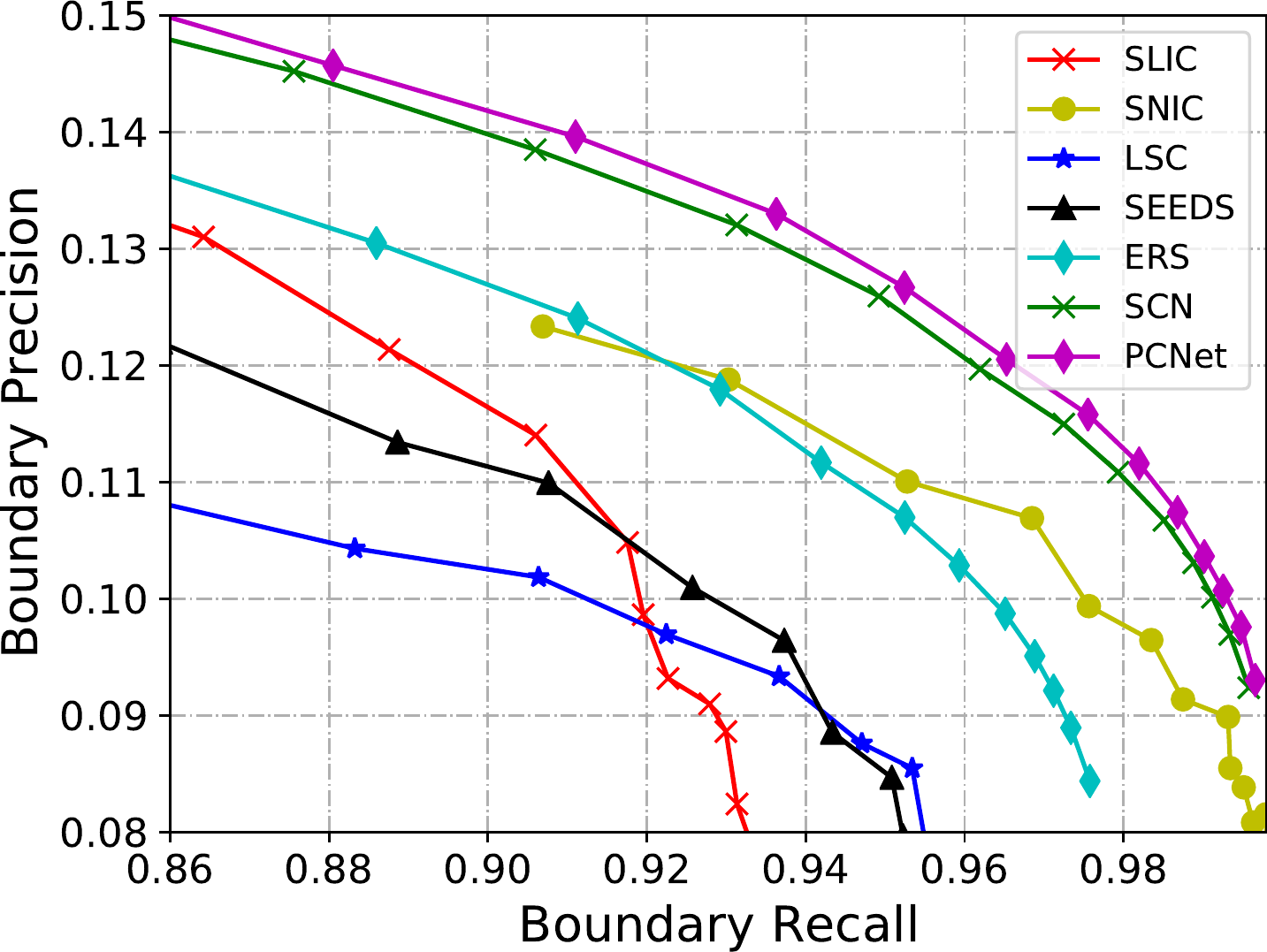}\\
        \vspace{-0.25cm}
        \mycaption{(a) Face-Human}
    \end{minipage}
    }
\subfigure{
    \begin{minipage}[t]{0.28\linewidth}
        \centering
        \includegraphics[width=2.1in]{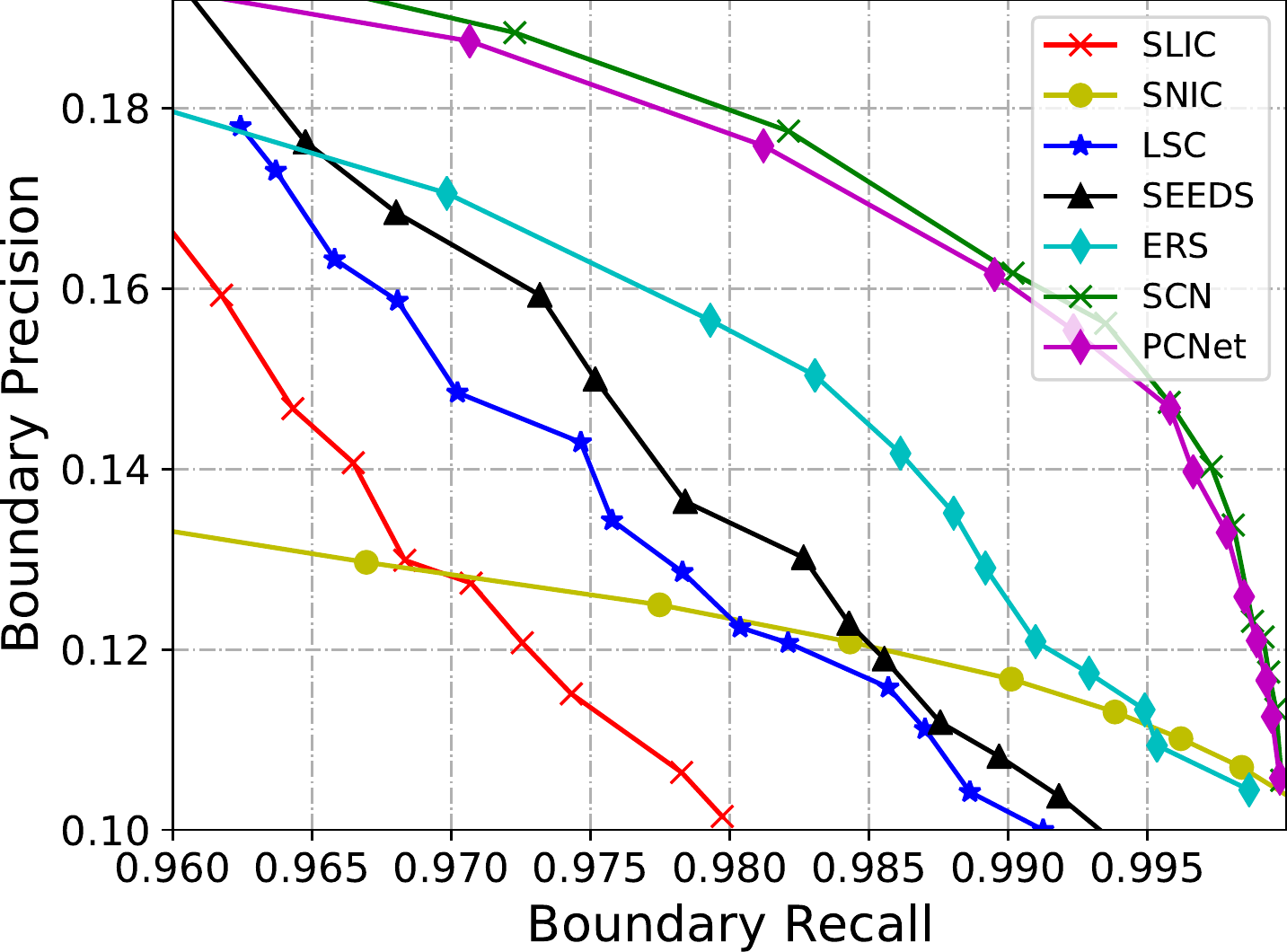}\\
        \label{rebuild:a}
        \vspace{-0.25cm}
        \mycaption{(b) Mapillary-Vistas}
    \end{minipage}
    }
\subfigure{
    \begin{minipage}[t]{0.28\linewidth}
        \centering
         \includegraphics[width=2.17in]{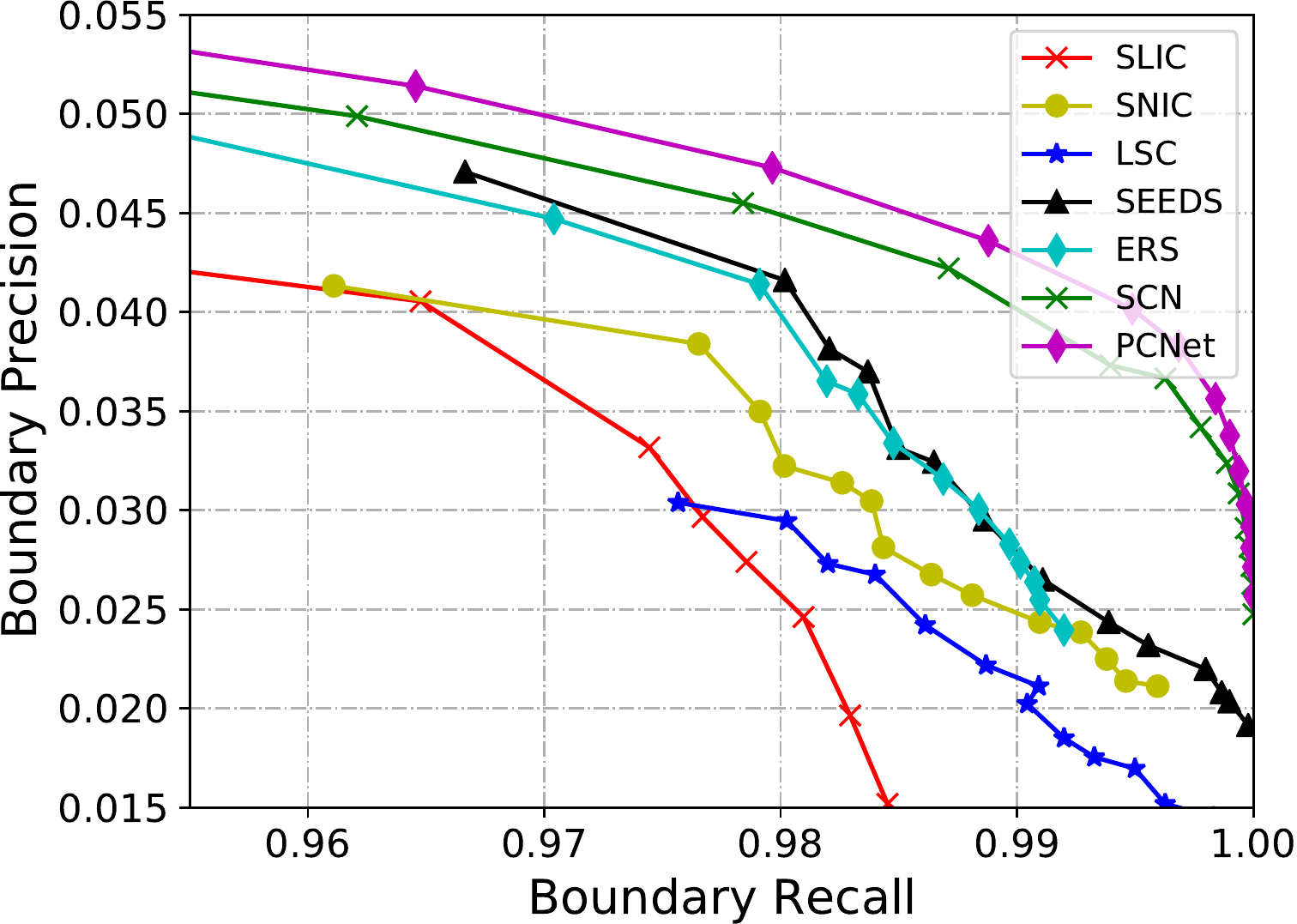}\\
           \vspace{-0.25cm}
        \mycaption{(c) BIG}
    \end{minipage}
    }
    \\
\centering\text{\normalsize I: trained on Face-Human.}\\
    \subfigure{
    \begin{minipage}[t]{0.28\linewidth}
        \centering
        \includegraphics[width=2.10in]{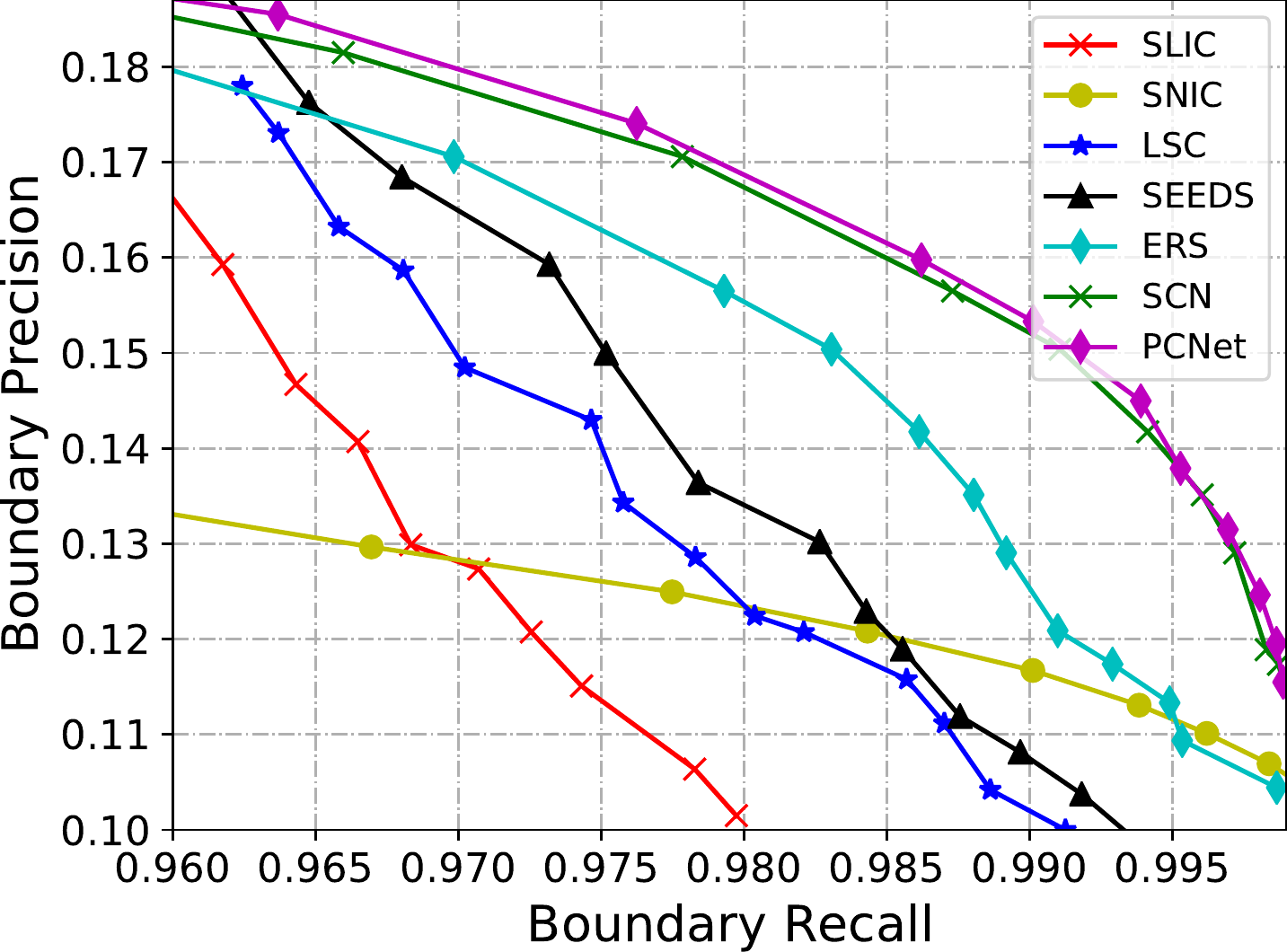}\\
        \vspace{-0.25cm}
        \mycaption{(a) Mapillary-Vistas}
    \end{minipage}
    }
\subfigure{
    \begin{minipage}[t]{0.28\linewidth}
        \centering
        \includegraphics[width=2.10in]{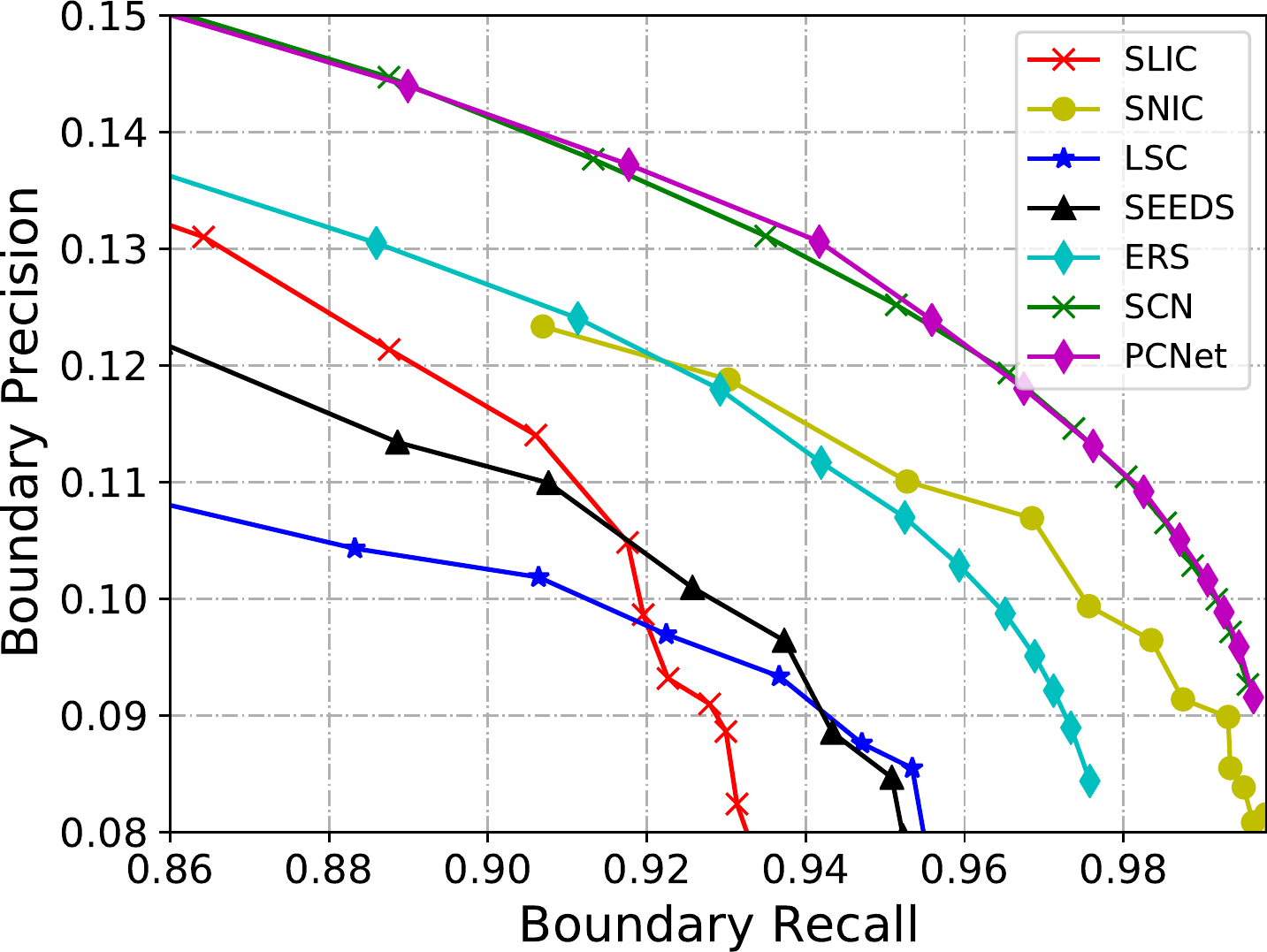}\\
        \label{rebuild:c}
        \vspace{-0.25cm}
        \mycaption{(b) Face-Human}
    \end{minipage}
    }
    \hspace{-0.05in}
   \subfigure{
    \begin{minipage}[t]{0.28\linewidth}
        \centering
        \includegraphics[height=1.60in,width=2.172in]{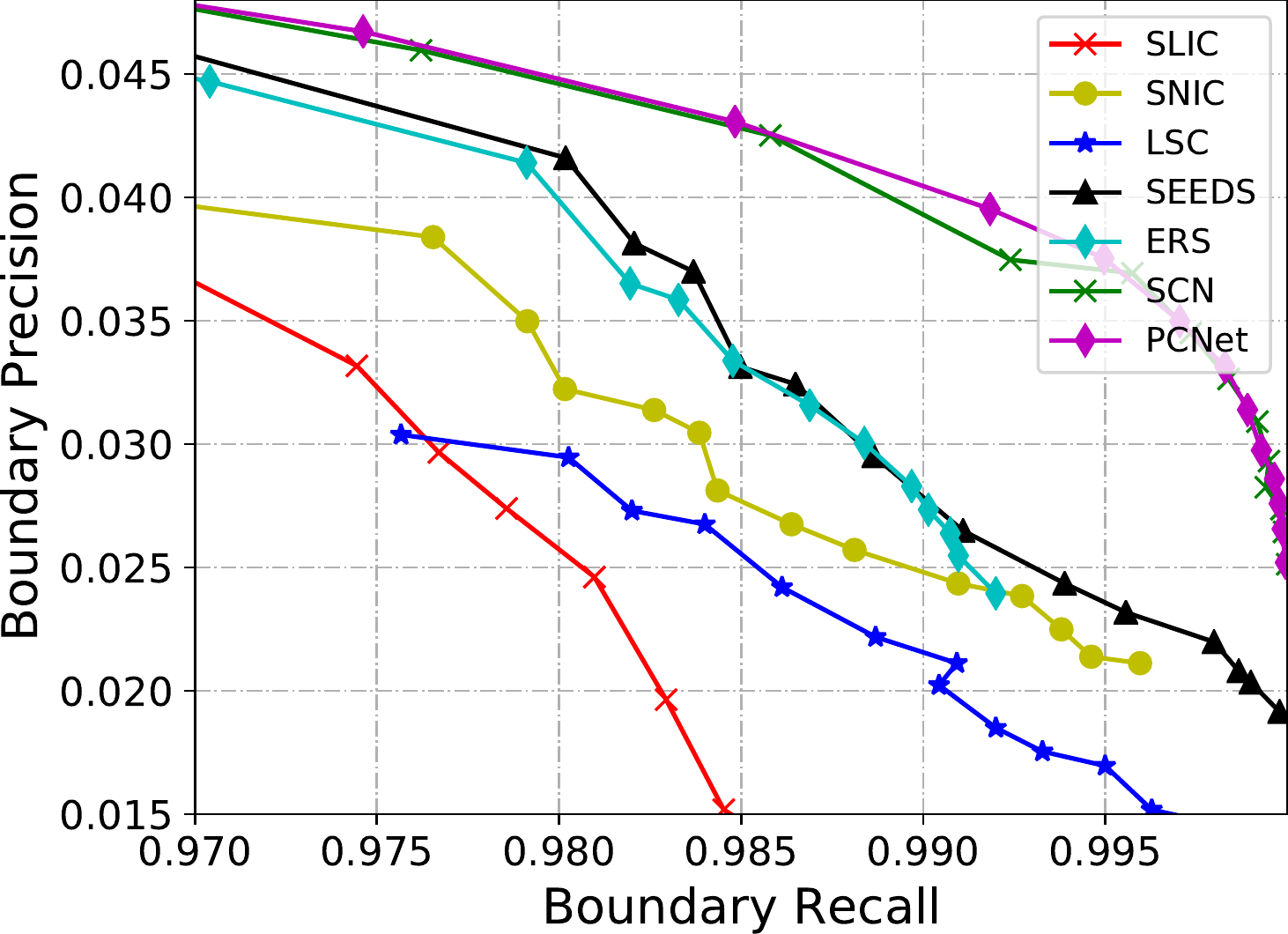}\\
        \label{rebuild:d}
        \vspace{-0.25cm}
        \mycaption{(c) BIG}
    \end{minipage}
    }
\centering\text{\normalsize II: trained on Mapillary-Vistas.}\\
\caption{The performance comparison on three high-resolution benchmarks. The top row shows the BR-BP curves of all models trained on Face-Human while evaluated on Face-Human, Mapillary-Vistas, and BIG datasets from left to right. And the bottom row analogously exhibits the performance comparison of all models trained on Mapillary-Vistas dataset.}
\vspace{-0.4cm}
\label{main_perform}
\end{center}
\end{figure*}

\begin{figure}[t]
\setlength{\abovecaptionskip}{-2pt} 
\begin{center}
    \subfigure{
    \begin{minipage}[t]{0.80\linewidth}
        \includegraphics[width=2.44in]{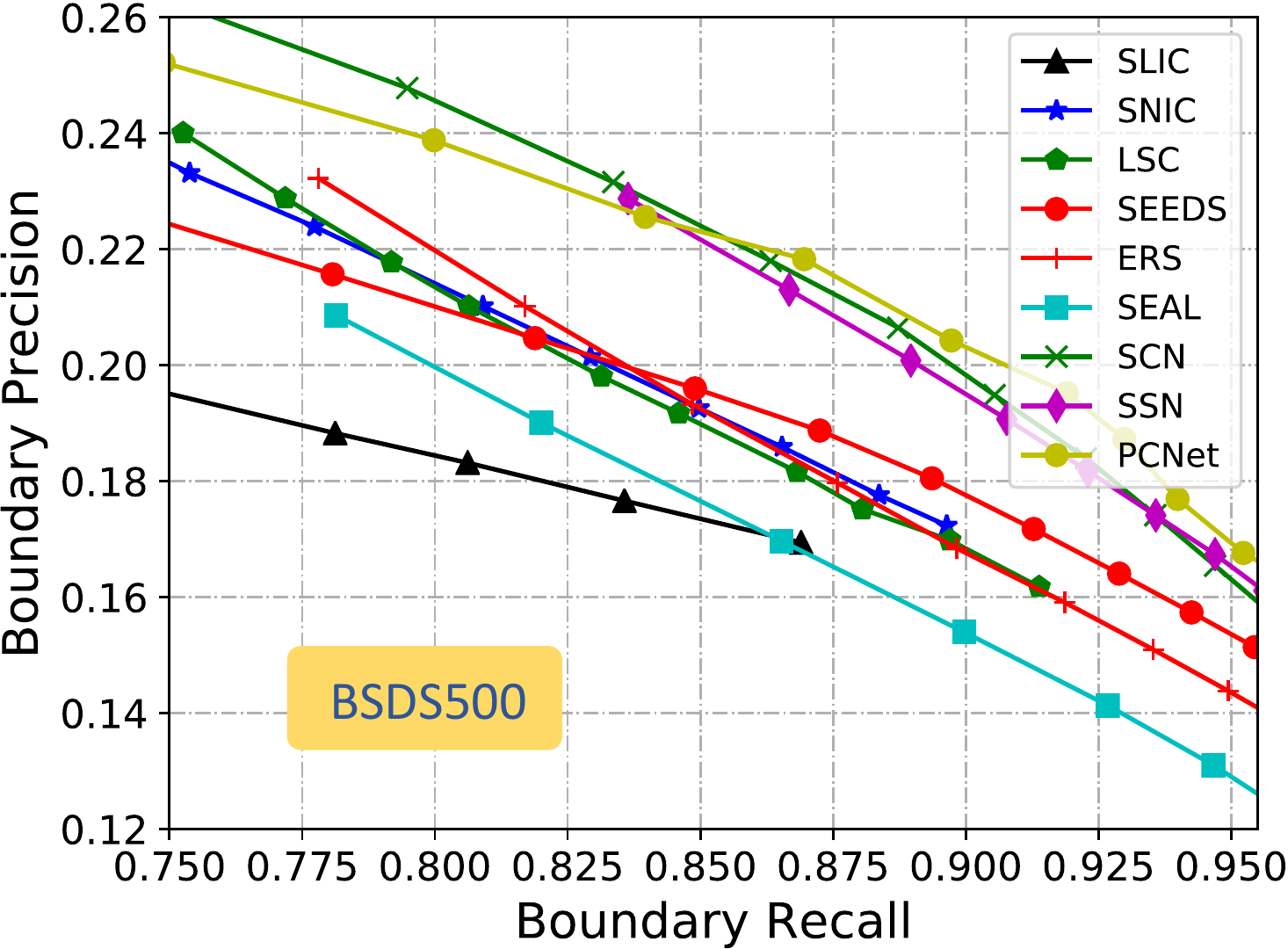}\\
    \vspace{-0.5cm}
    \end{minipage}
    }
    \\
    \vspace{-0.2cm}
    \subfigure{\begin{minipage}[t]{0.80\linewidth}
    \includegraphics[width=2.44in]{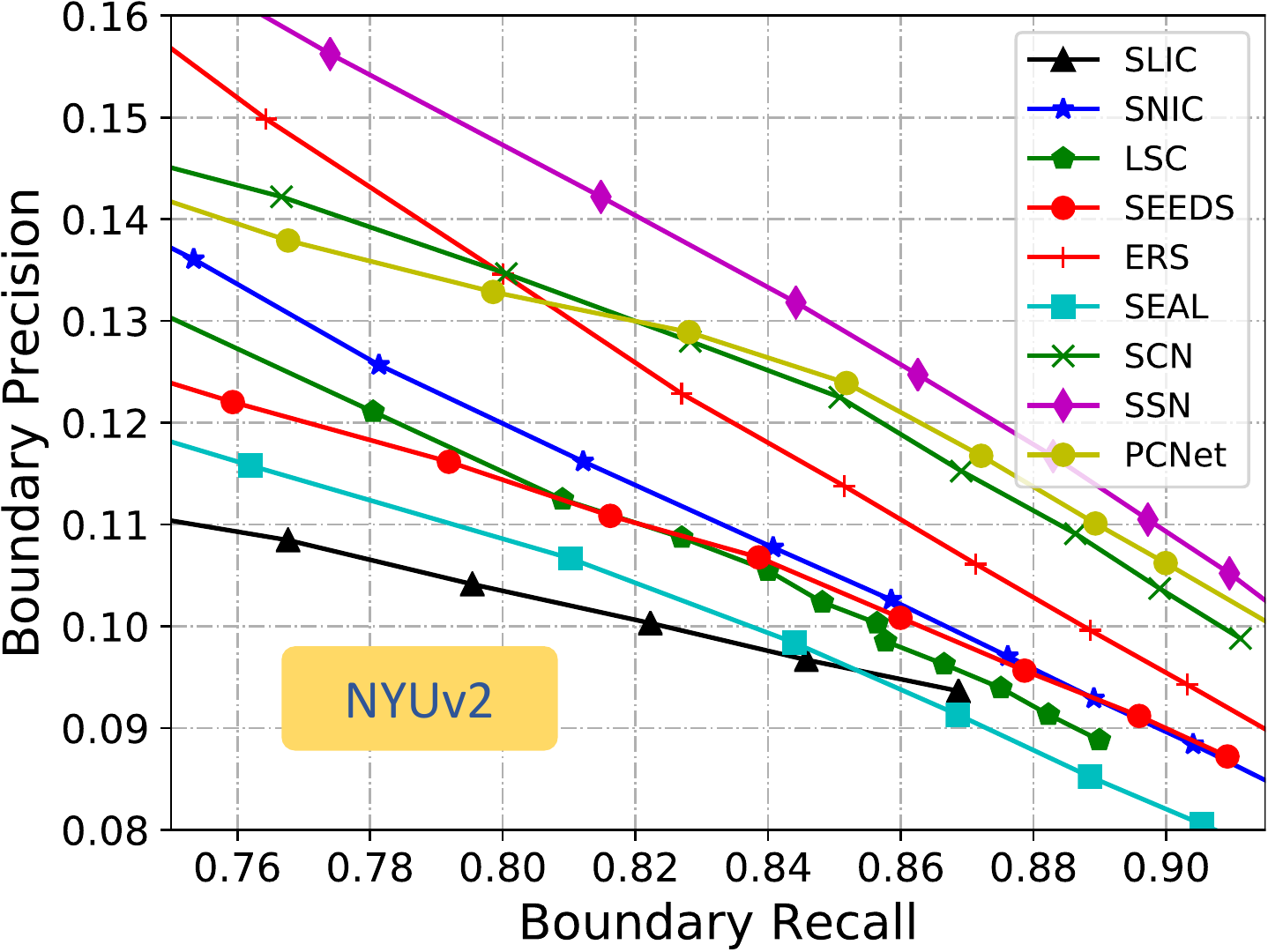}\\
    \vspace{-0.5cm}
    \end{minipage}}
\end{center}
\vspace{-0.2cm}
\caption{The performance comparison on popular datasets BSDS500 and NYUv2}
\label{BDS_NYU}
\end{figure}

\begin{figure*}[t]
\begin{center}
    \subfigure[Inputs]{
    \begin{minipage}[t]{0.14\linewidth}
    \includegraphics[width=1.06in]{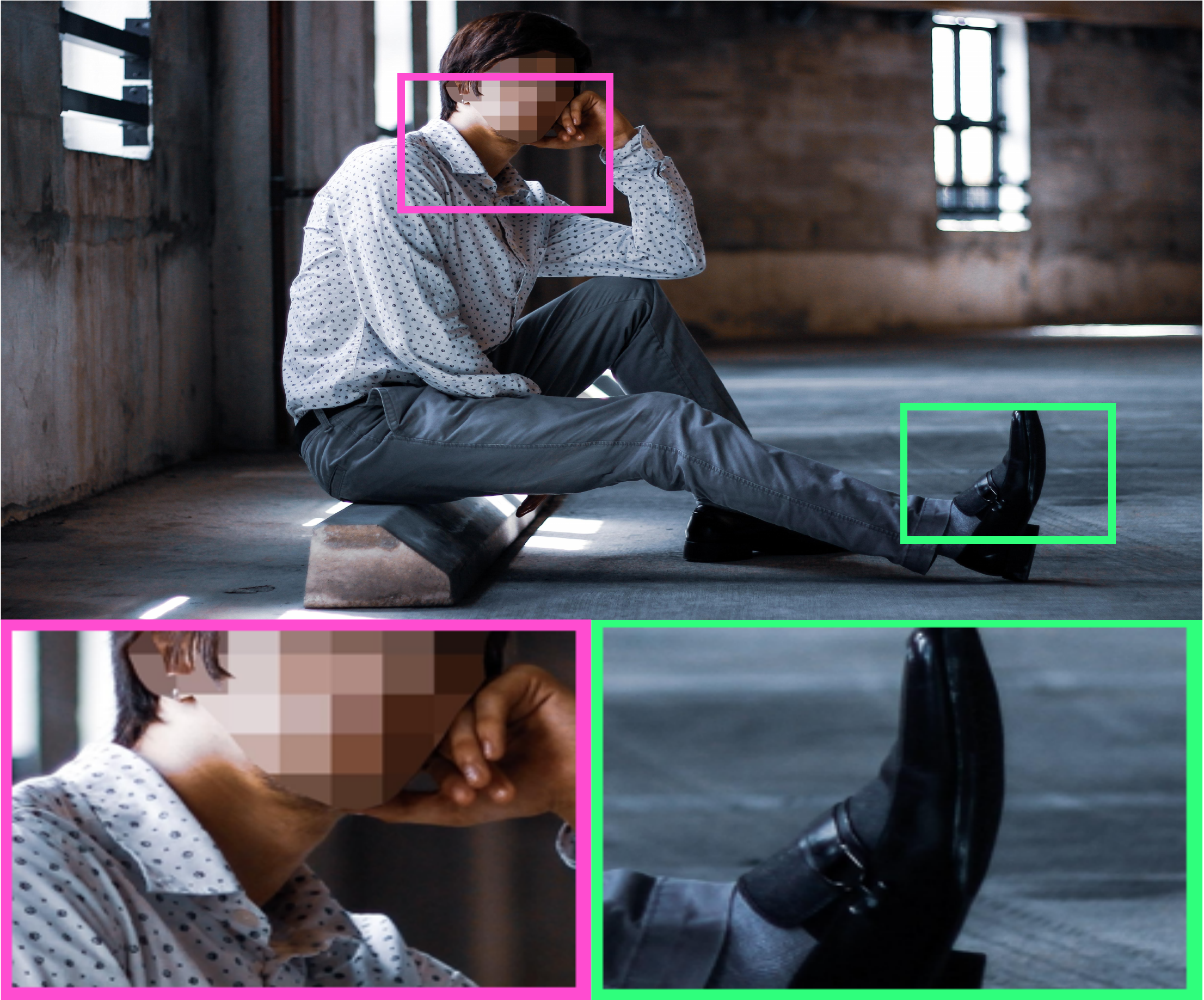}\\
    \vspace{-0.3cm}
    \includegraphics[width=1.06in]{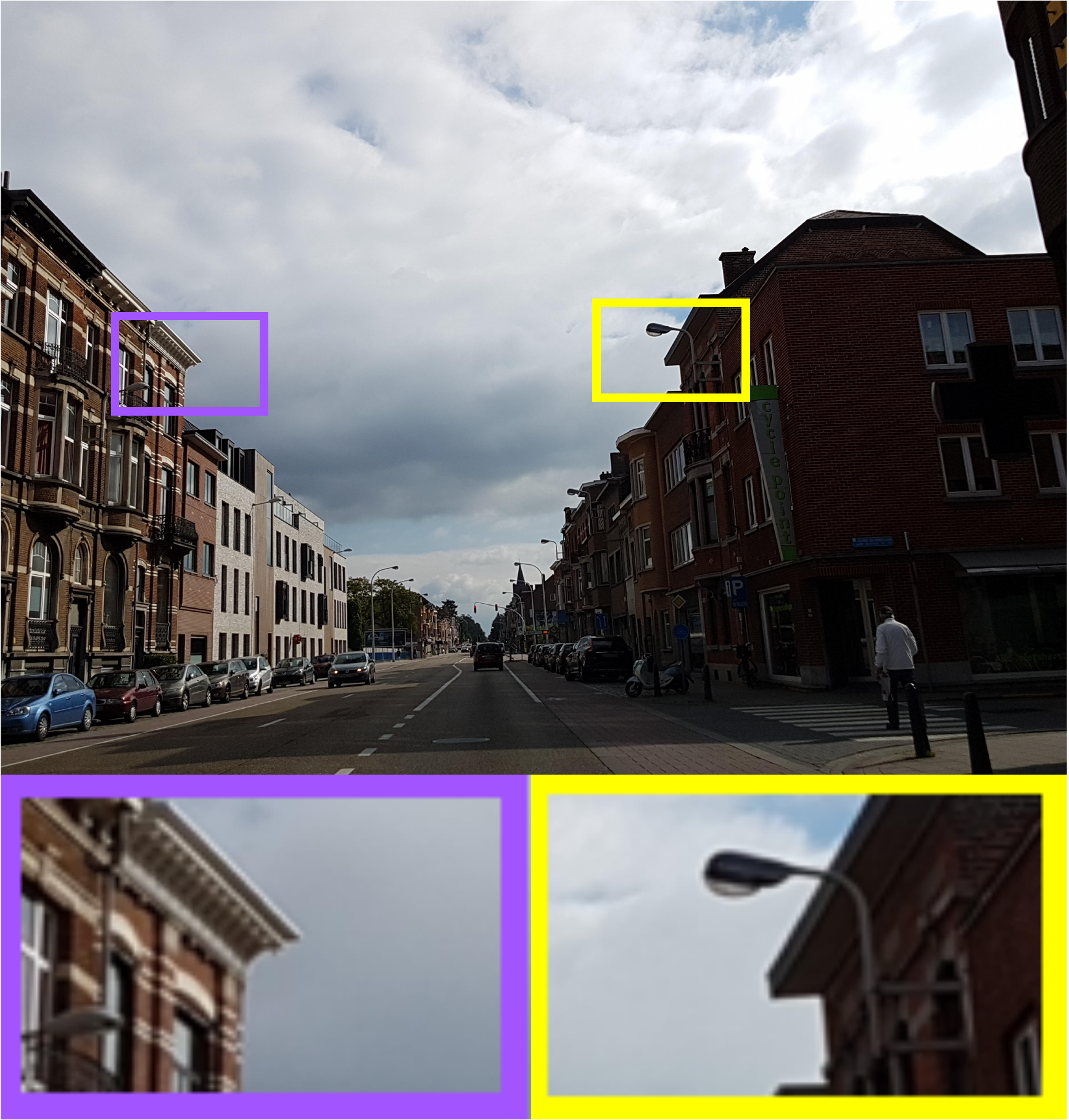}\\
     \vspace{-0.3cm}
    \includegraphics[width=1.06in]{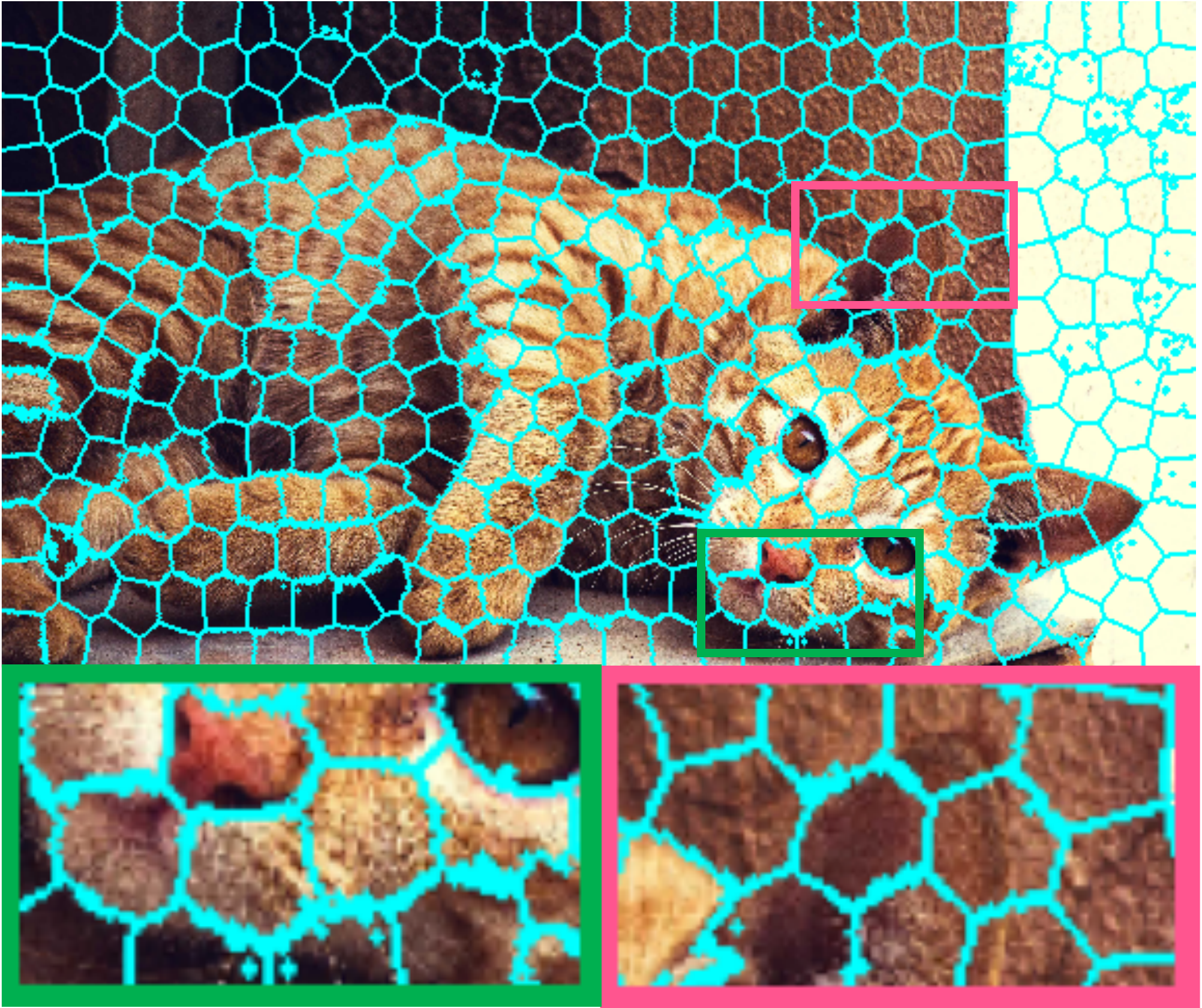}\\
    \vspace{-0.3cm}
    \end{minipage}
    }
   \hspace{-0.04cm}
    \subfigure[GT labels]{
    \begin{minipage}[t]{0.15\linewidth}
    \includegraphics[width=1.06in]{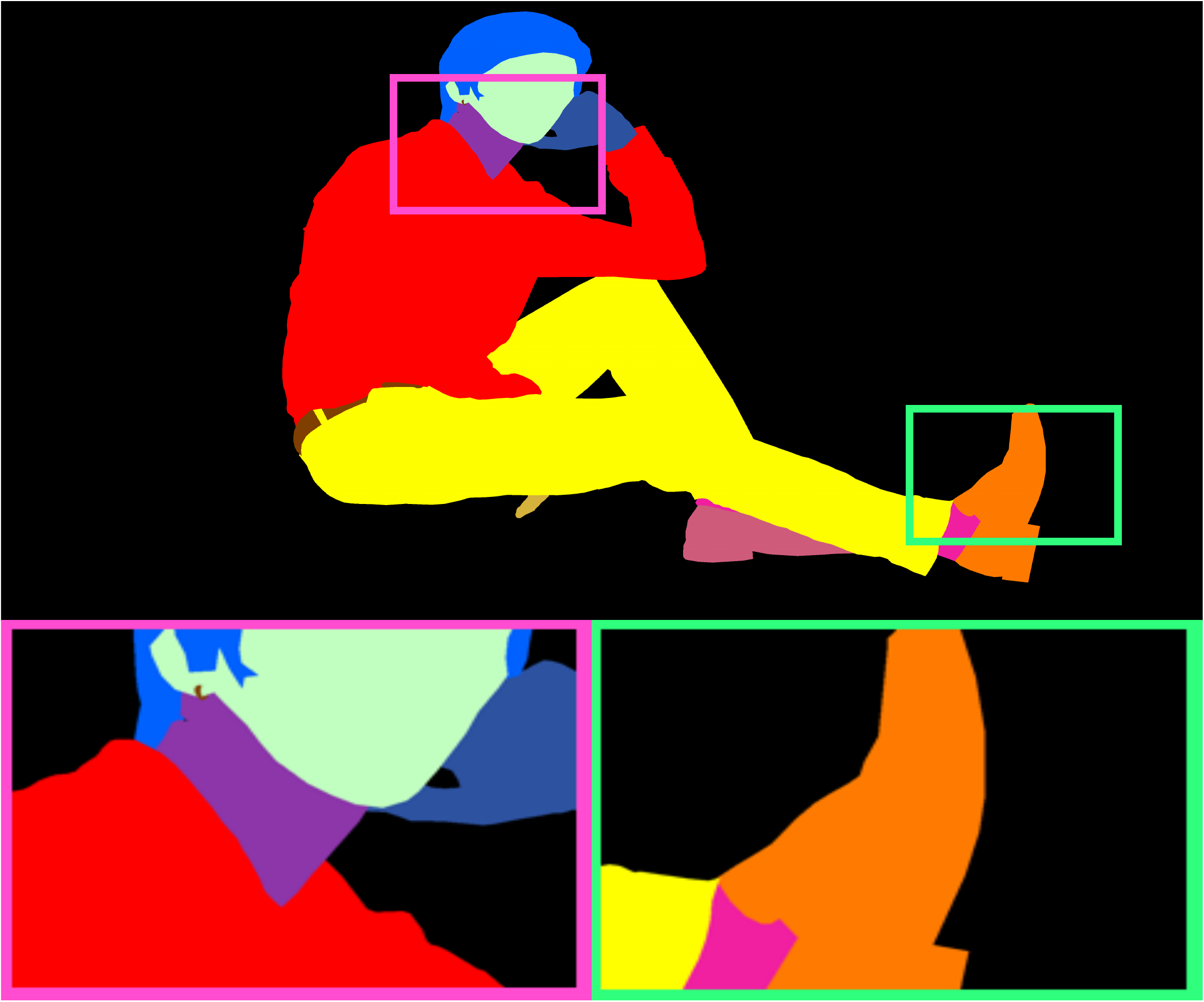}\\
     \vspace{-0.3cm}
    \includegraphics[width=1.06in]{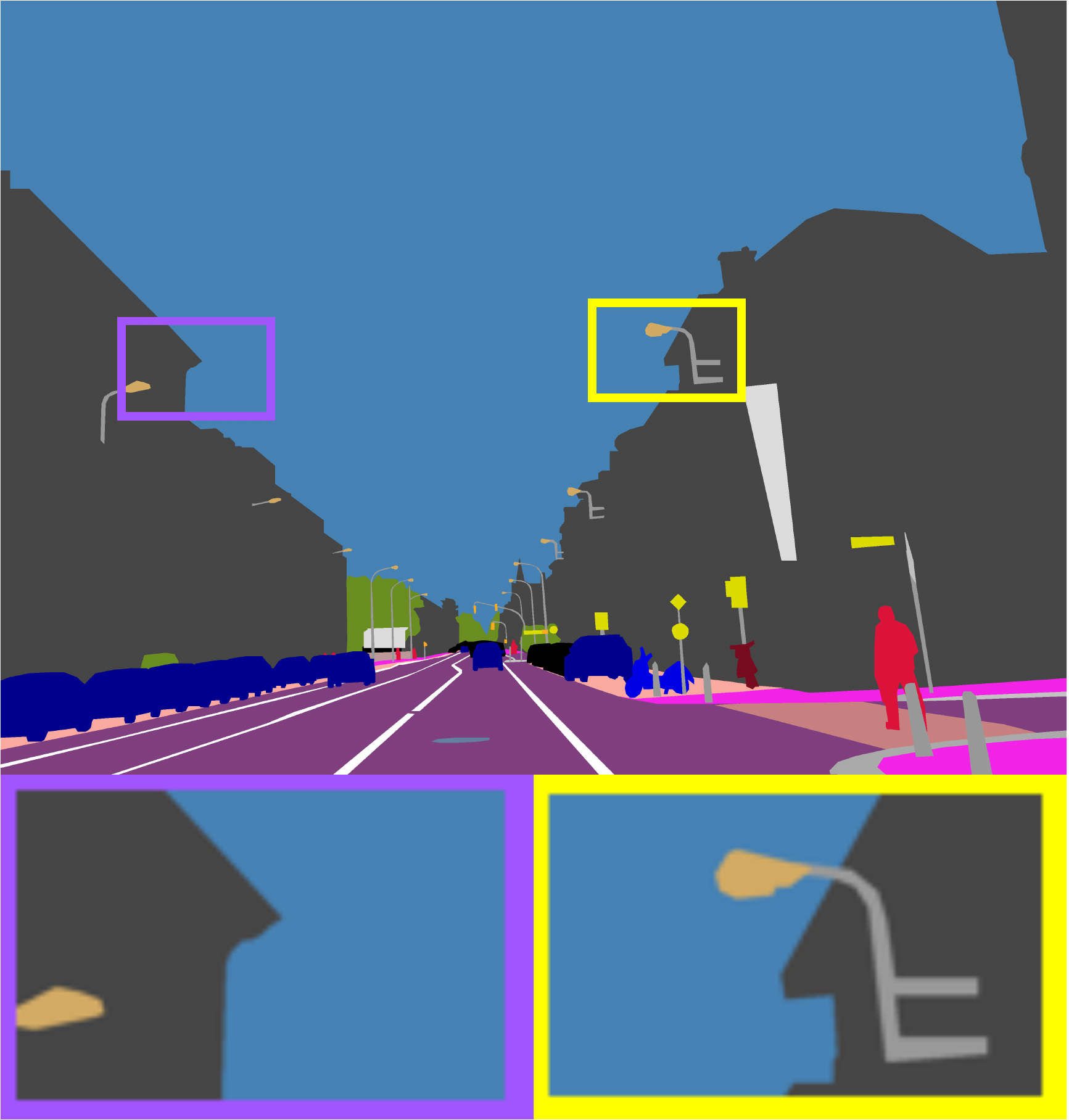}\\
     \vspace{-0.3cm}
    \includegraphics[width=1.06in]{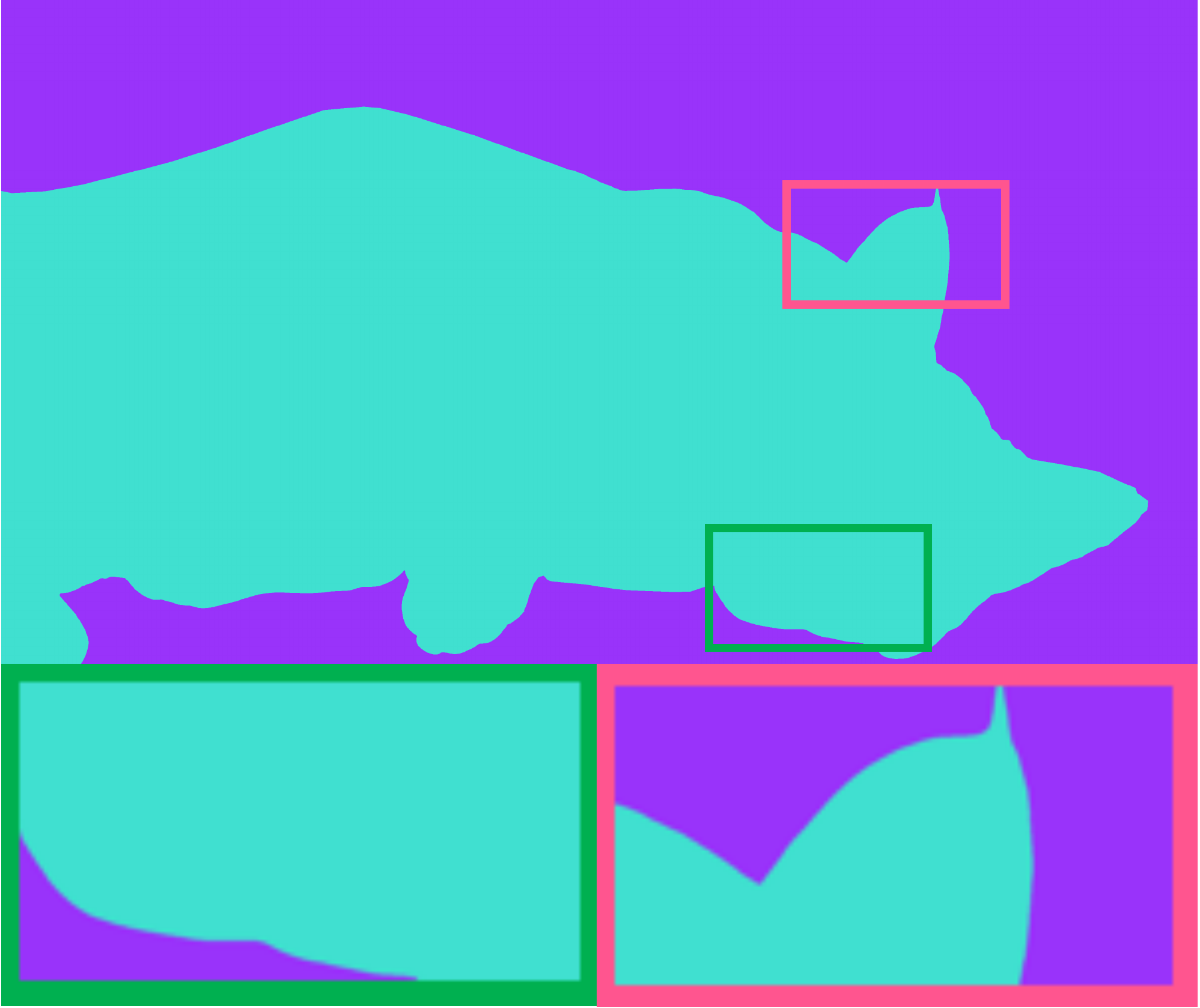}\\
     \vspace{-0.3cm}
    \end{minipage}}
    \hspace{-0.06cm}
    \subfigure[SLIC~\cite{SLIC}]{
    \begin{minipage}[t]{0.15\linewidth}
    \includegraphics[width=1.06in]{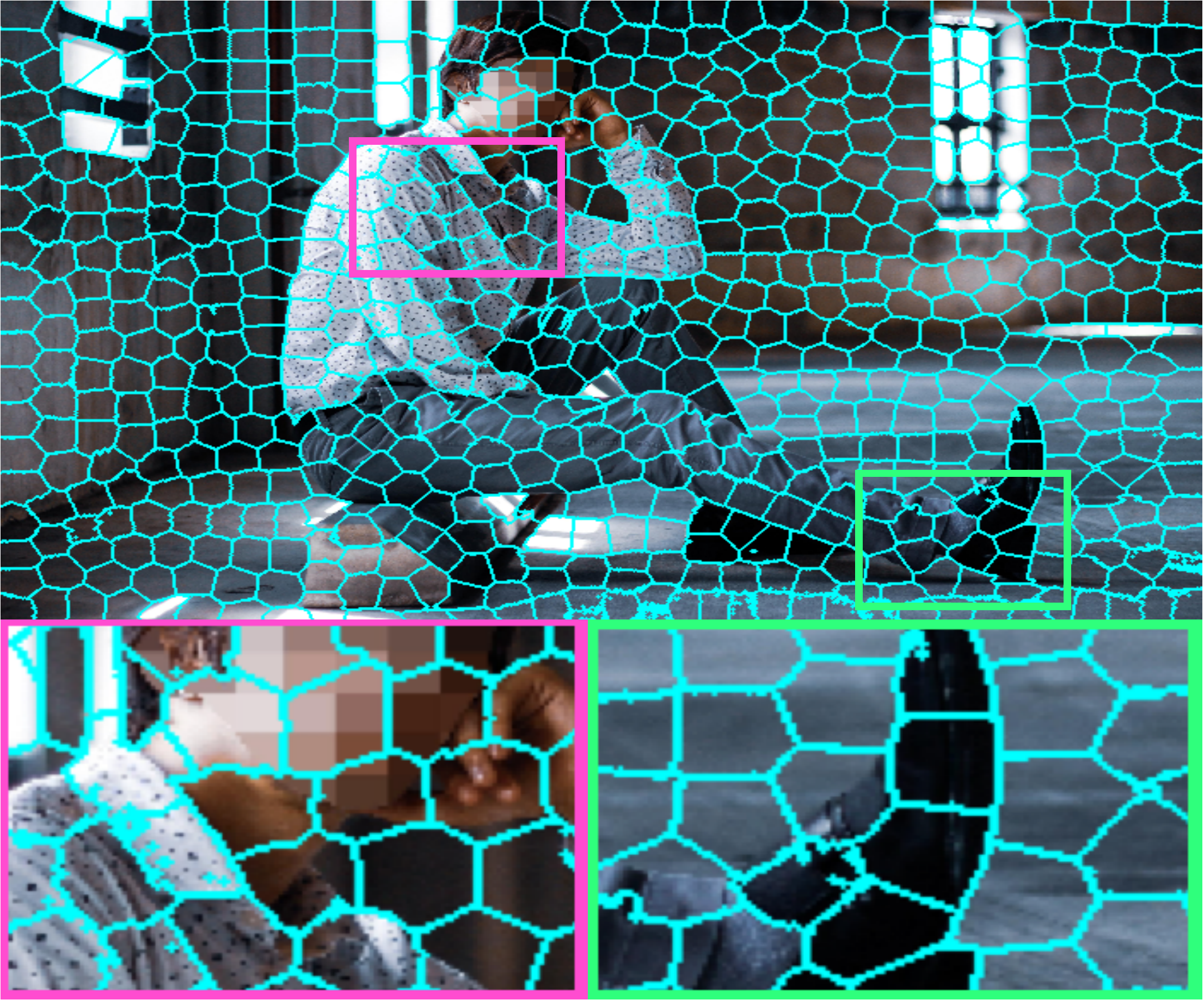}\\
     \vspace{-0.3cm}
    \includegraphics[width=1.06in]{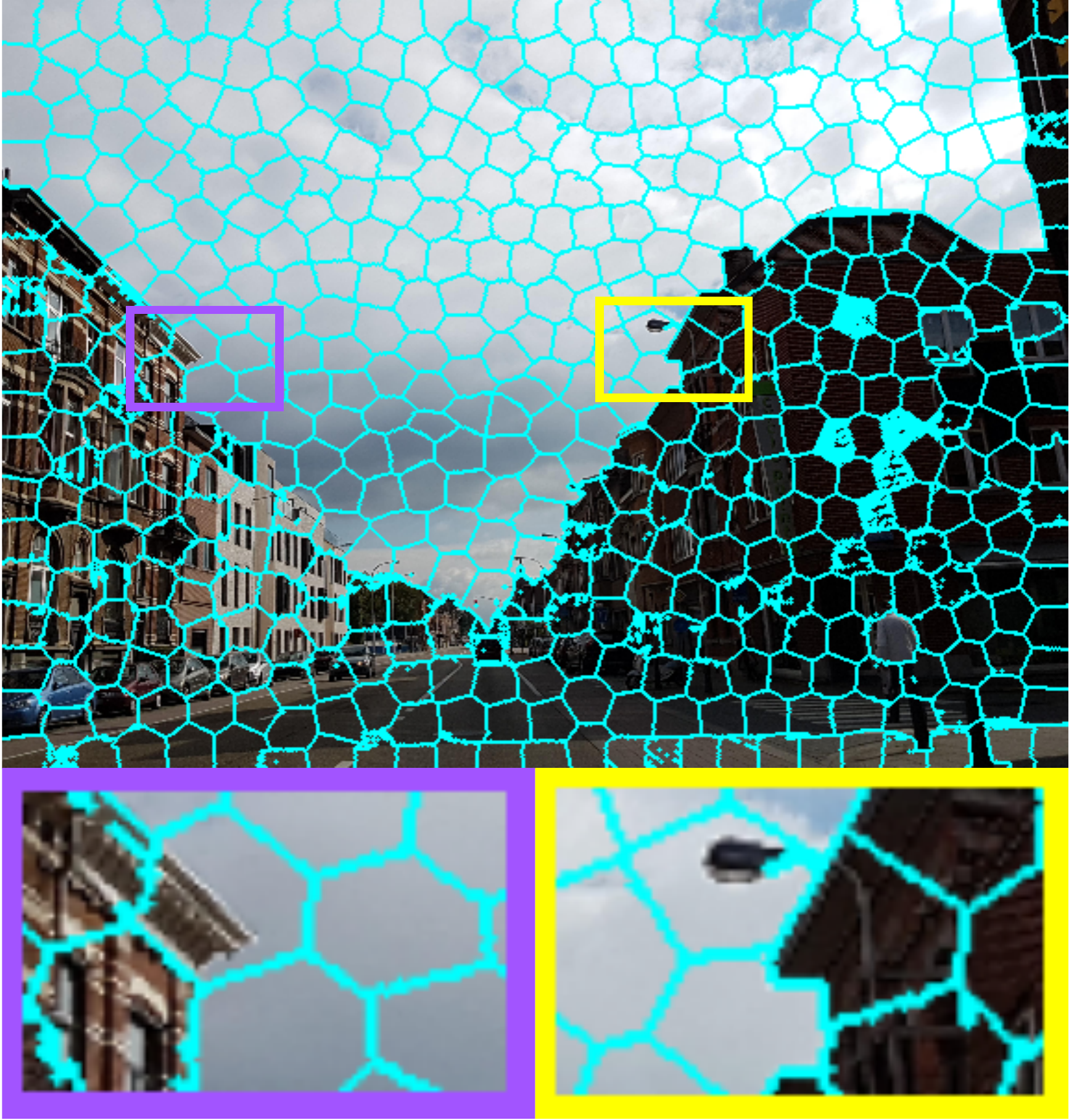}\\
     \vspace{-0.3cm}
    \includegraphics[width=1.06in]{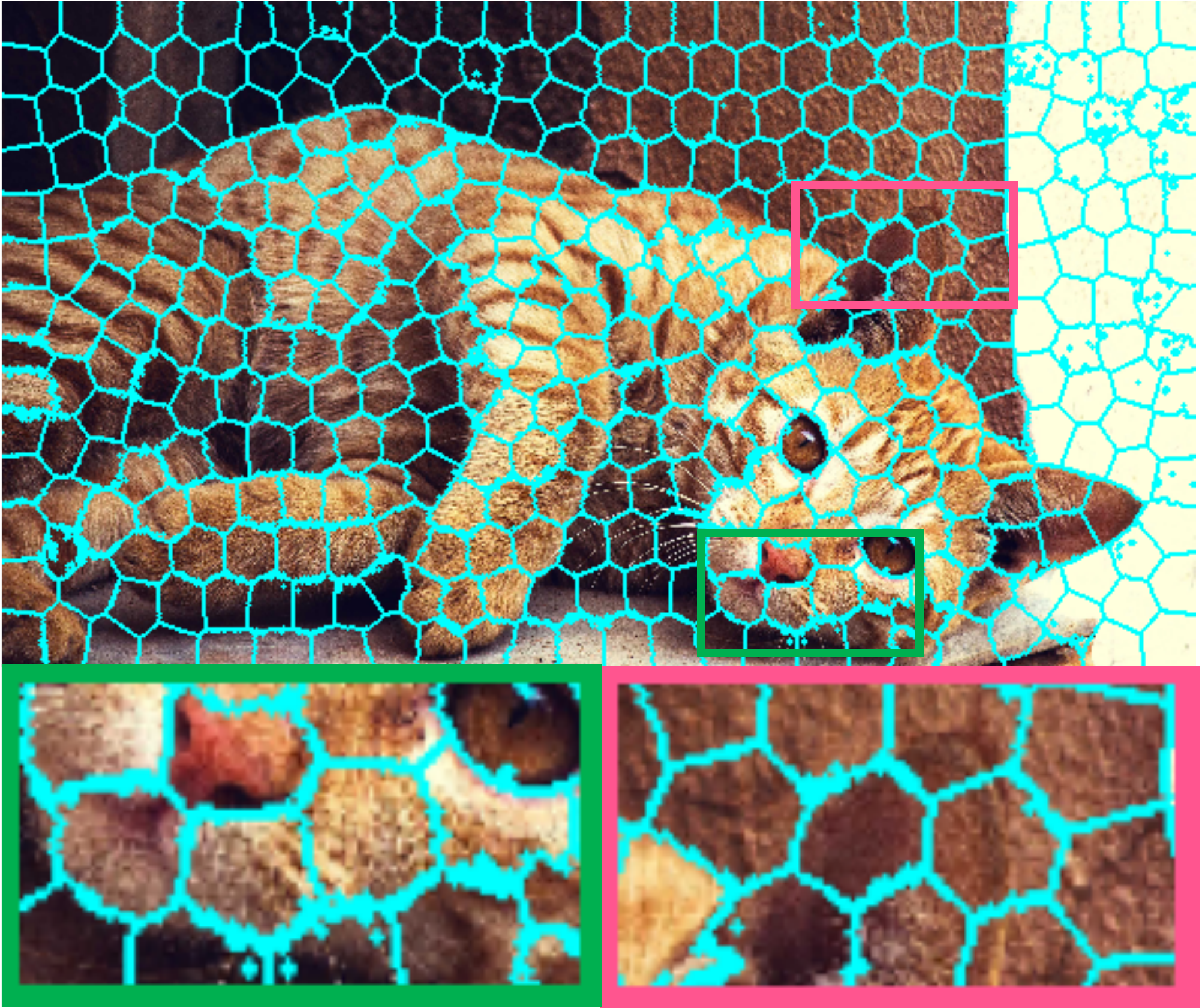}\\
     \vspace{-0.3cm}
    \end{minipage}
    }
    \hspace{-0.18cm}
    \subfigure[SNIC~\cite{SNIC}]{
    \begin{minipage}[t]{0.15\linewidth}
    \includegraphics[width=1.06in]{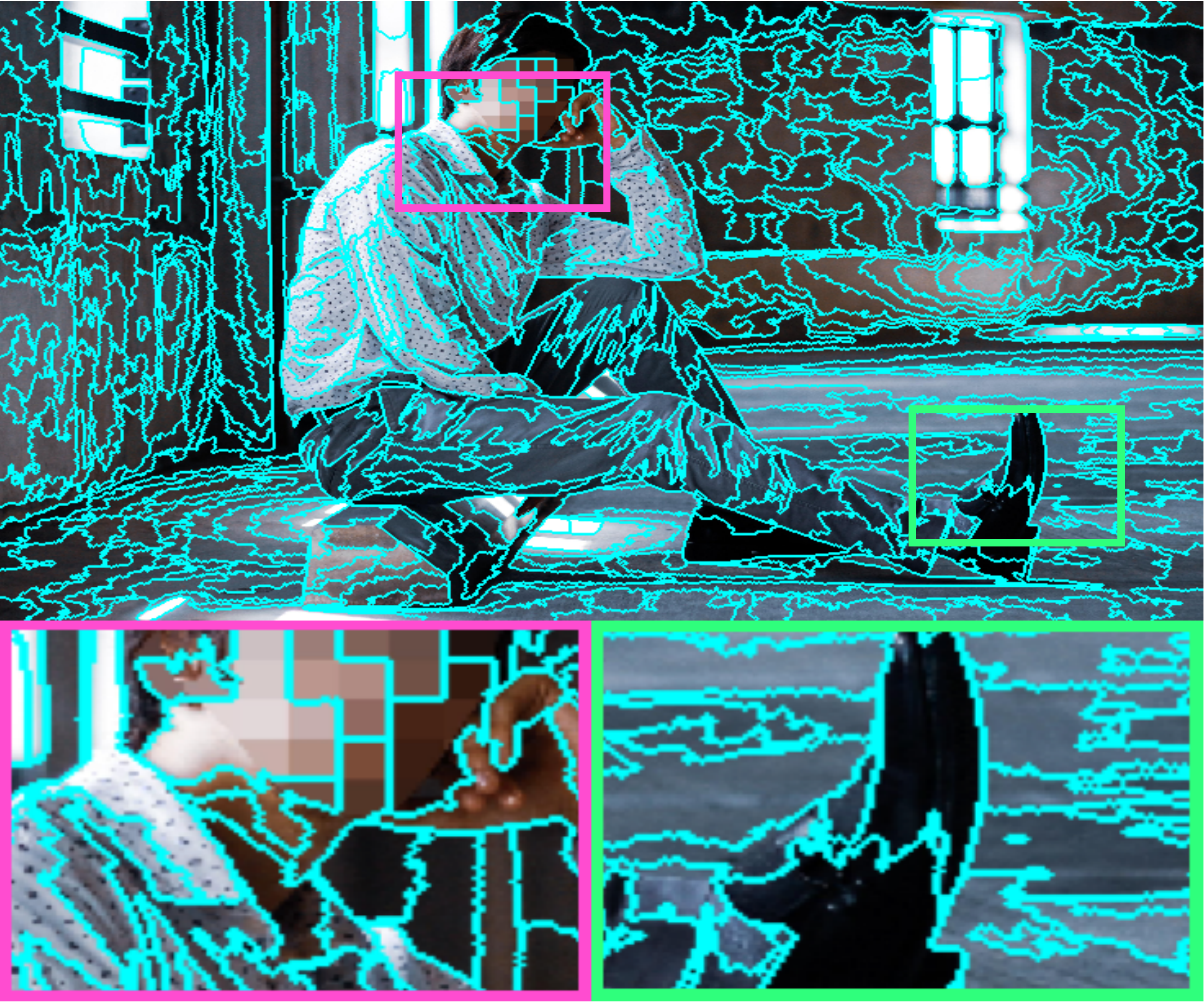}\\
     \vspace{-0.3cm}
    \includegraphics[width=1.06in]{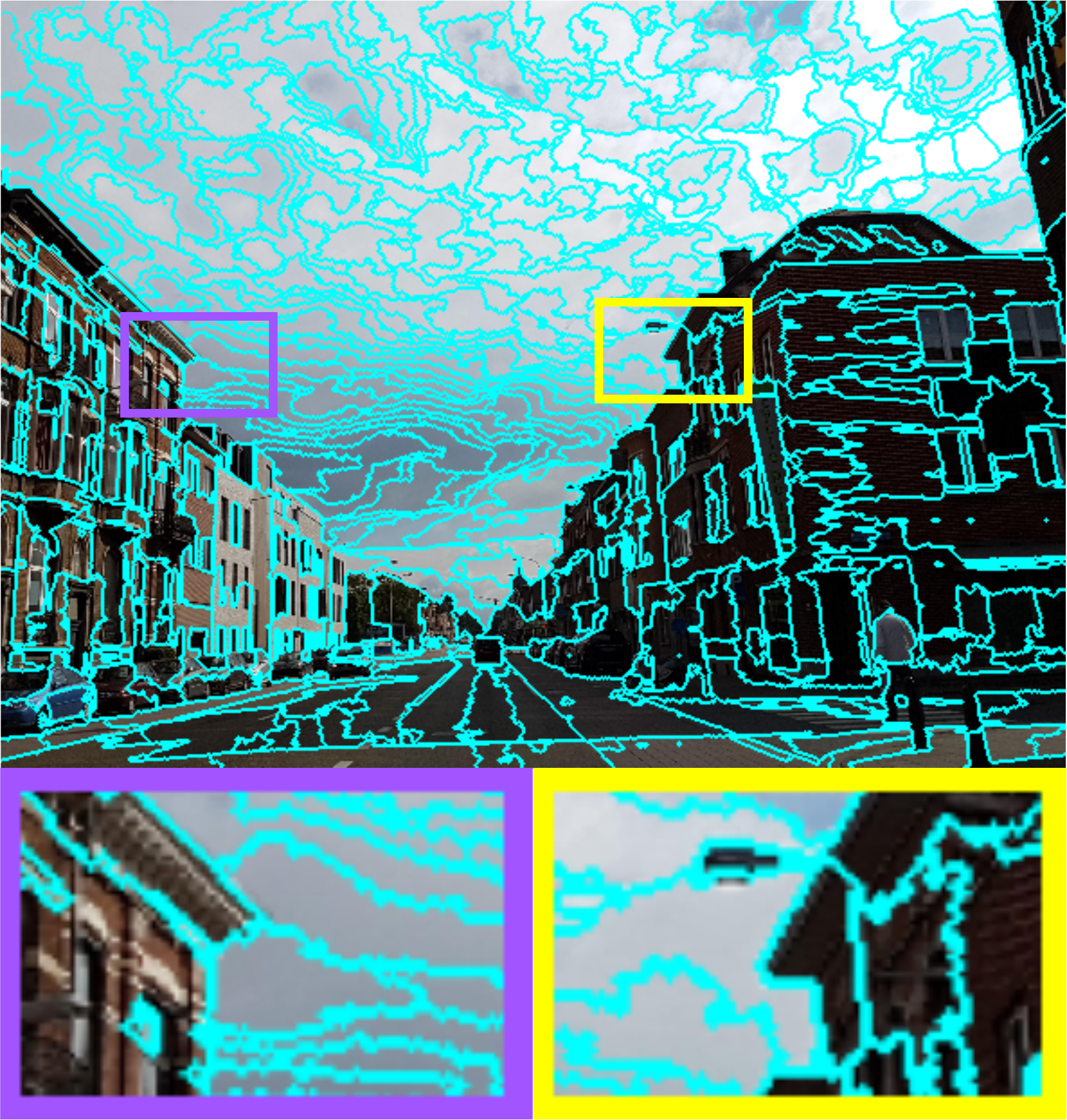}\\
     \vspace{-0.3cm}
    \includegraphics[width=1.06in]{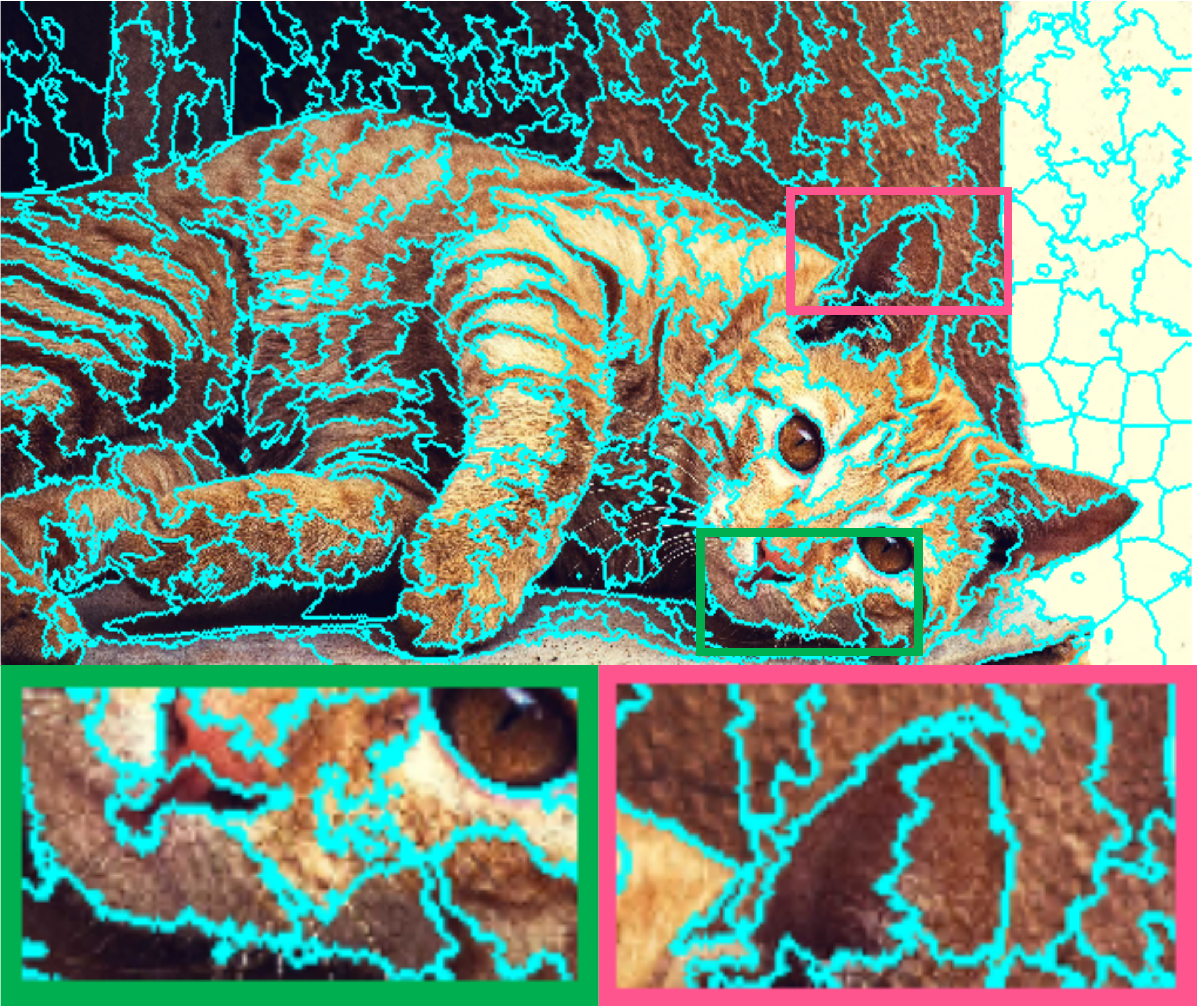}\\
     \vspace{-0.3cm}
    \end{minipage}
    }
    \hspace{-0.18cm}
    \subfigure[SCN~\cite{SCN}]{
    \begin{minipage}[t]{0.15\linewidth}
    \includegraphics[width=1.06in]{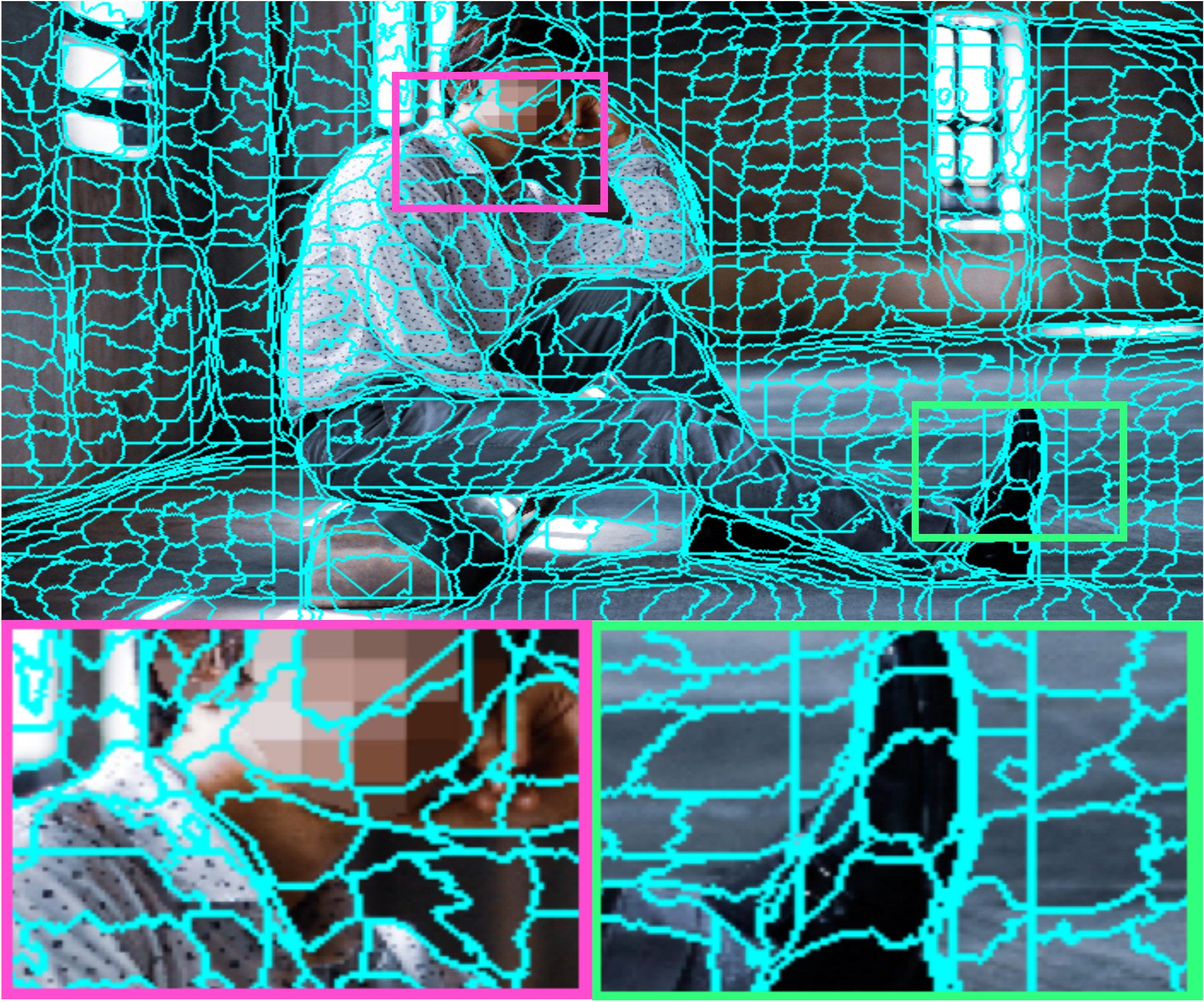}\\
    \vspace{-0.3cm}
    \includegraphics[width=1.06in]{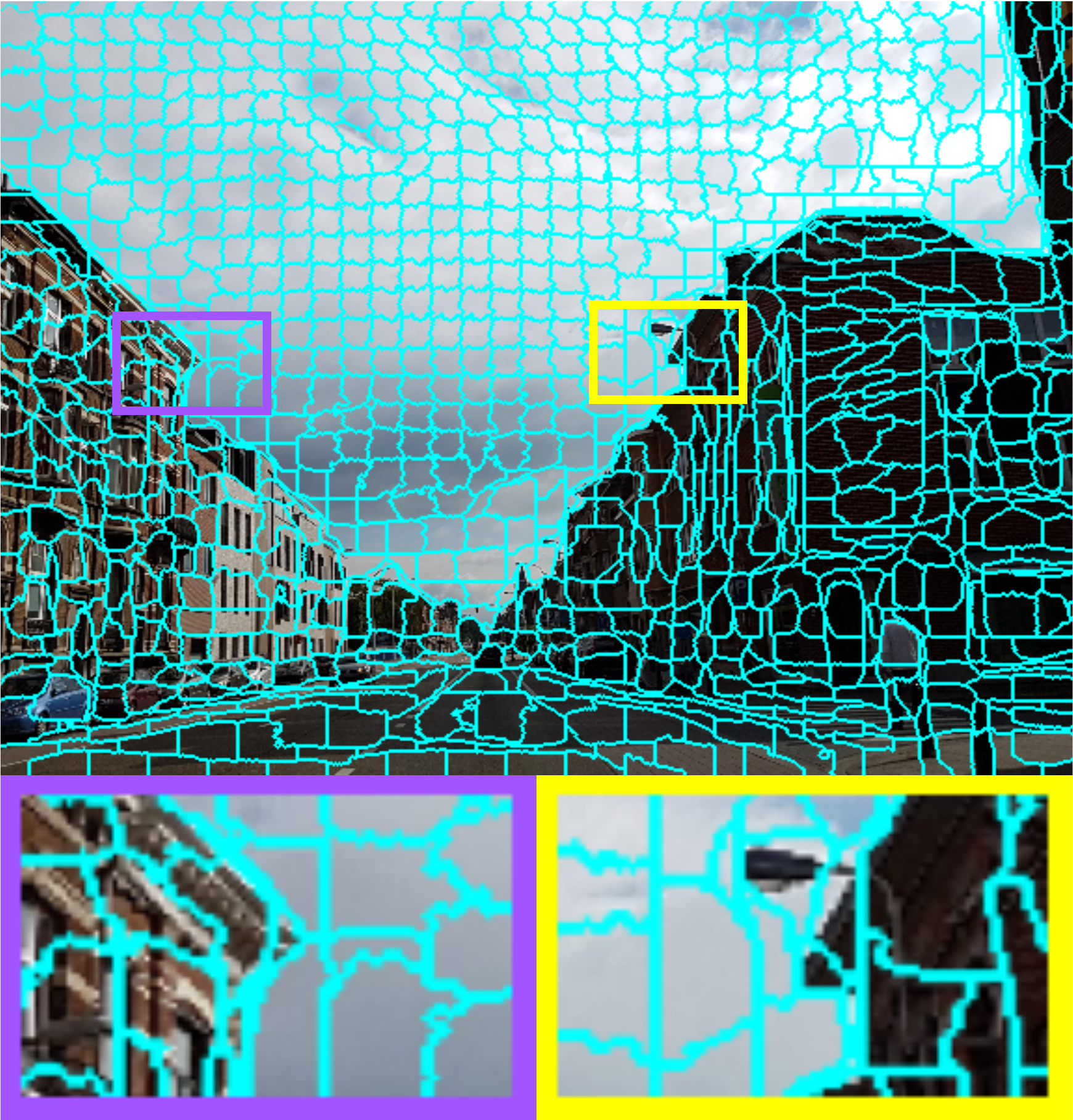}\\
     \vspace{-0.3cm}
    \includegraphics[width=1.06in]{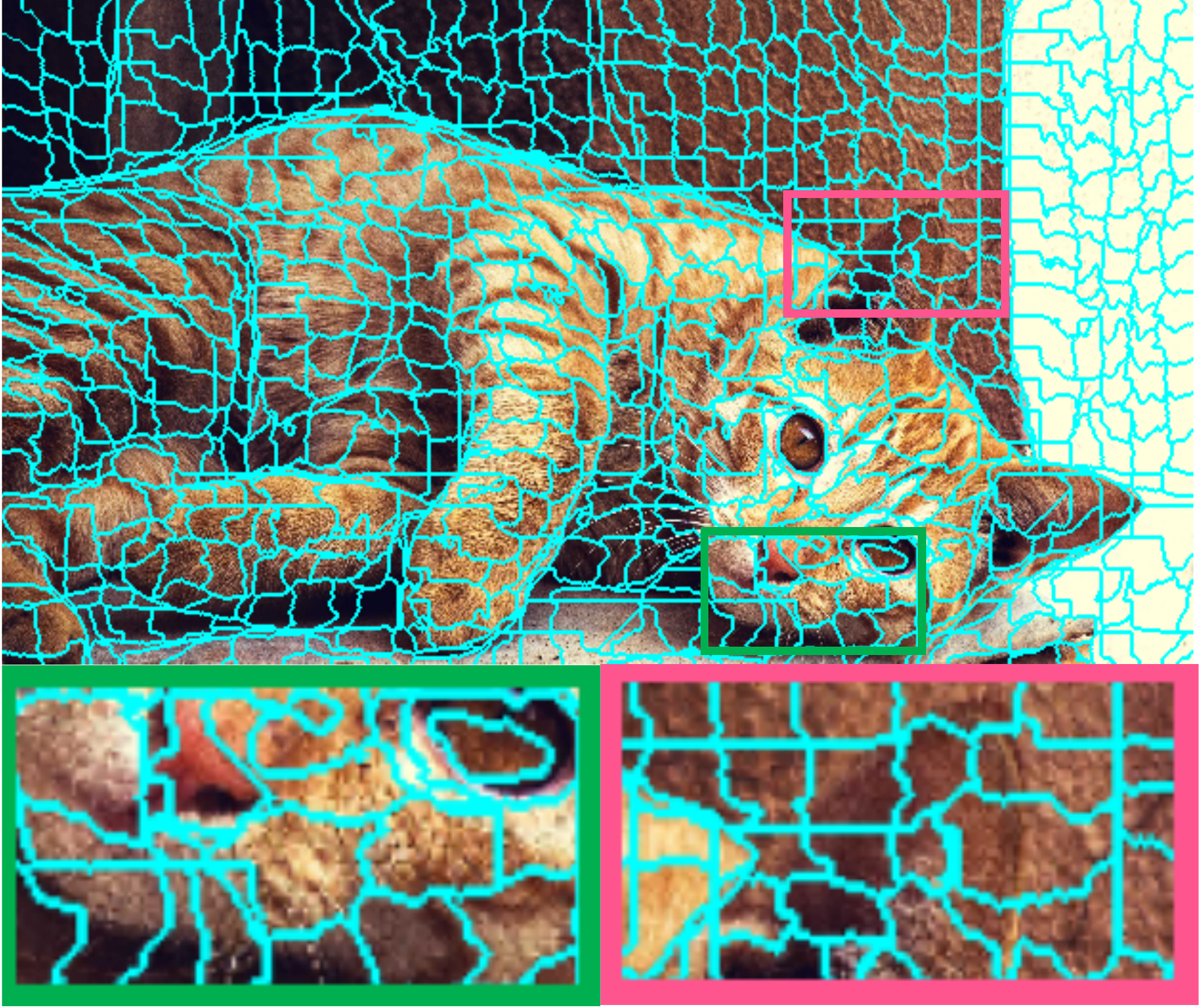}\\
     \vspace{-0.3cm}
    \end{minipage}
    }
     \hspace{-0.18cm}
    \subfigure[PCNet]{
    \begin{minipage}[t]{0.15\linewidth}
    \includegraphics[width=1.06in]{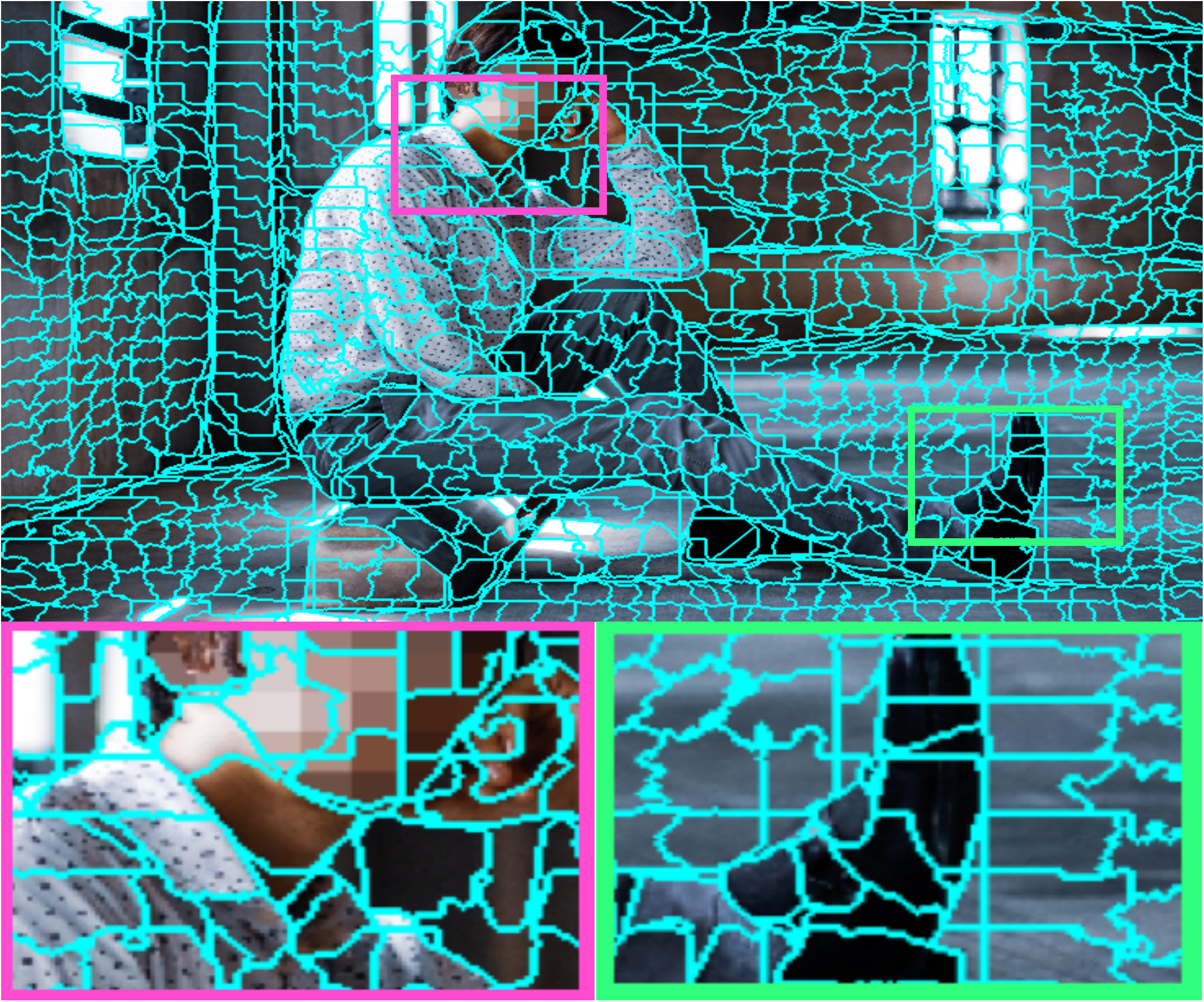}\\
    \vspace{-0.3cm}
    \includegraphics[width=1.06in]{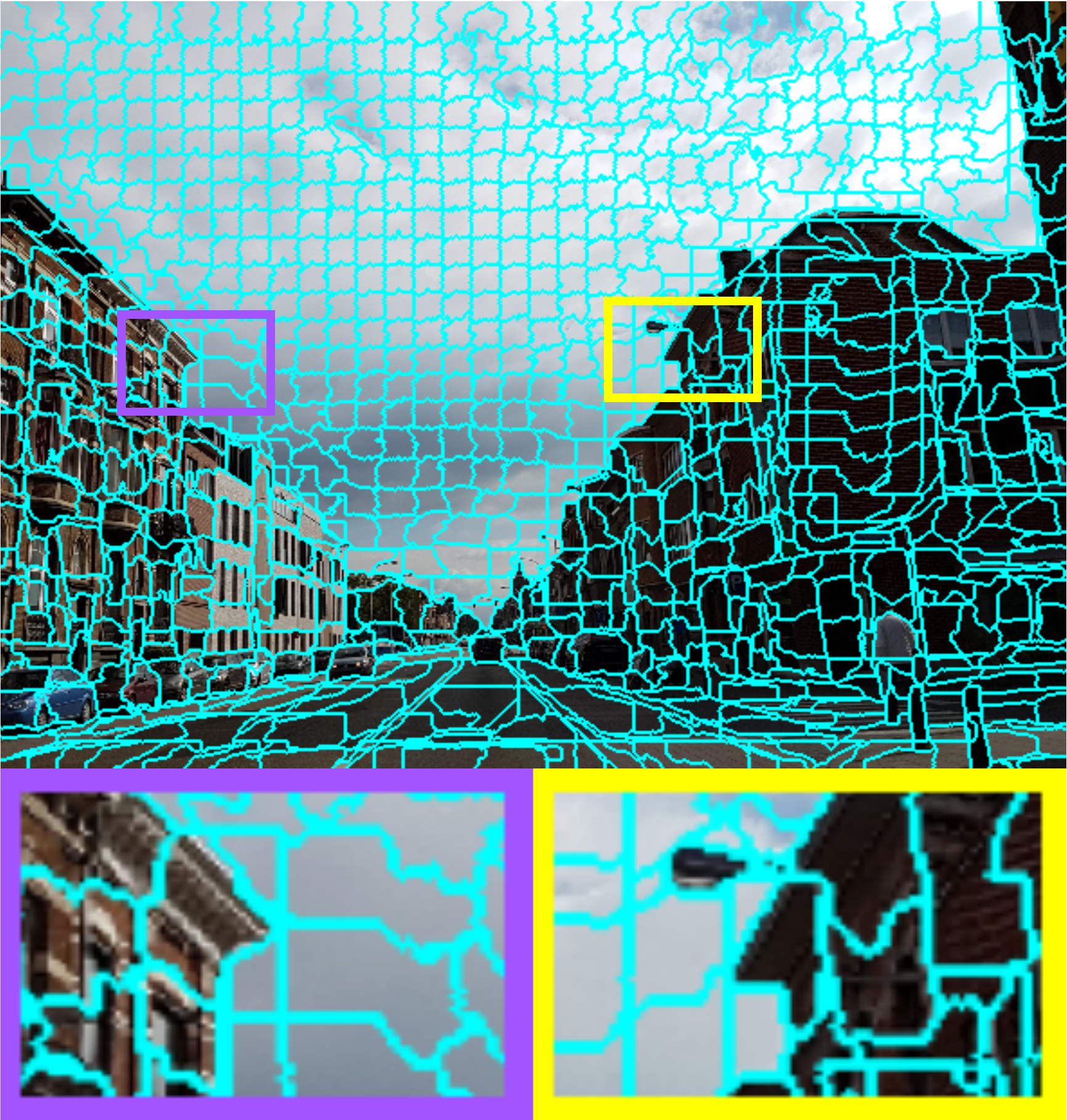}\\
     \vspace{-0.3cm}
    \includegraphics[width=1.06in]{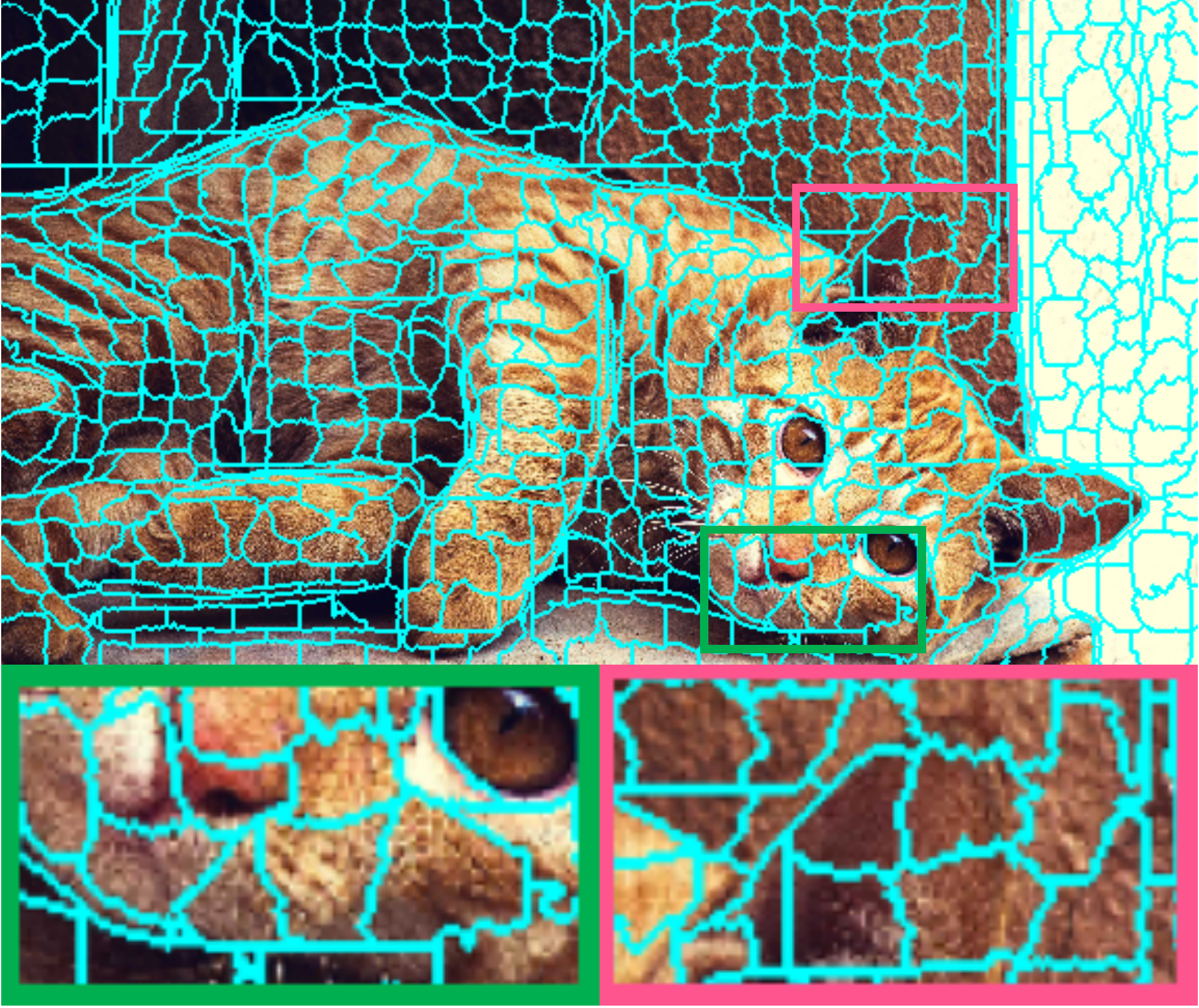}\\
     \vspace{-0.3cm}
    \end{minipage}
    }
\end{center}
\vspace{-0.5cm}
 \caption{Qualitative results of four superpixel methods, SLIC, SNIC, SCN, and our PCNet. The top to bottom rows subsequently  exhibits the results of Face-Human, Mapillary-Vistas, and BIG datasets.}
 \vspace{0.2cm}
\label{main_visual}
\end{figure*}

\begin{figure}[t]
\setlength{\abovecaptionskip}{-2pt} 
\begin{center}
    \subfigure[2$\times$2 division]{
    \begin{minipage}[t]{0.47\linewidth}
        \includegraphics[width=1.74in]{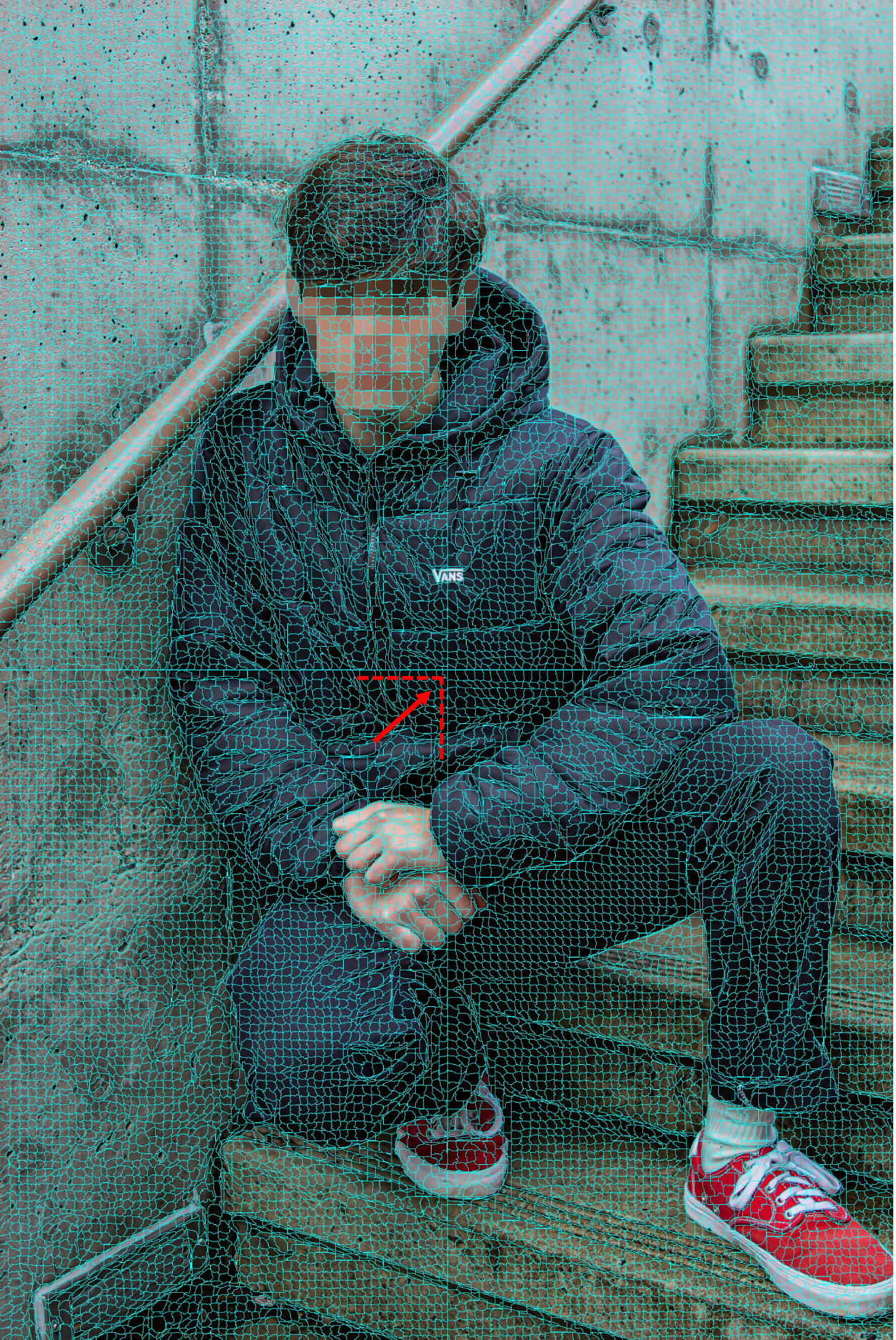}\\
    \vspace{-0.15cm}
    \end{minipage}
    }
    \hspace{0.01cm}
    \subfigure[4$\times$4 division]{\begin{minipage}[t]{0.47\linewidth}
    \includegraphics[width=1.74in]{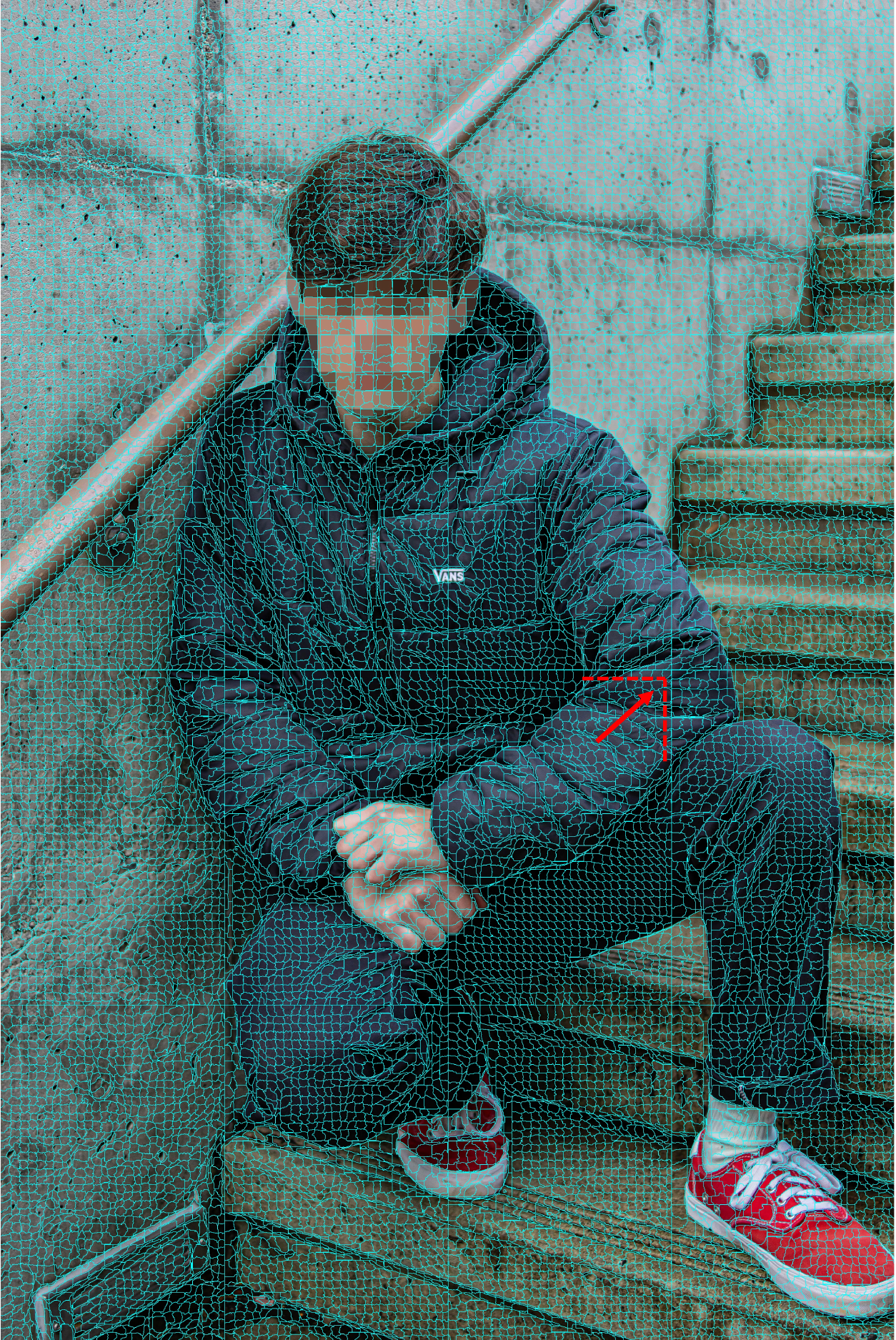}\\
    \vspace{-0.15cm}
    \end{minipage}}
\end{center}
\vspace{-0.15cm}
 \caption{The visual results of division-and-conquer strategy with 2$\times$2 division and 4$\times$4 division. The dashed lines indicate the boundaries between different image patches.}
 \vspace{0.6cm}
\label{dc_show}
\end{figure}

Overall, our full training loss is formulated as:
\begin{equation}
    \label{full_loss}
    \begin{aligned}
    \mathcal{L} =\mathcal{L}_G + \alpha\mathcal{L}_L +\beta\mathcal{L_D},
    \end{aligned}
\end{equation}
where $\alpha, \beta\in\mathcal{R}$ are trade-off hyperparameters.  The training procedure of our PCNet is summarized in Algorithm \textcolor{red}{1}. When inference, the super-resolution branch is discarded.

\section{Experiments}
\label{experiment}
We conduct extensive experiments on five datasets including three high-resolution datasets and two common benchmarks whose samples are with regular size. We systematically make comparison with state-of-the-art and classic methods to evaluate the performance of our PCNet. 
\subsection{Datasets}
To thoroughly validate  the efficiency of our proposed method, we collect a very high-resolution dataset, named \textbf{Face-Human}. The Face-Human dataset comprises 250 face and 250 human images in total, the size of images ranges from $1810\times 2066$ to $10000\times 6675$.
All samples are carefully given the pixel-wise annotations by experts, specifically, the face images are manually labelled as 21 classes, while the human samples are assigned 24 labels. Fig.~\ref{data_show} gives two examples of the image and the corresponding label format. Our collected Face-Human dataset covers a large interval resolution and is  challenging enough to evaluate the very high-resolution superpixel and  image segmentation. In our experiments,  300 images are adopted for training, 80 for validation and 120 for test purpose. 

Furthermore, we also randomly sample a subset of \textbf{Mapillary Vistas}~\cite{vistas}, which contains 600 samples with resolution ranging from  $1024\times 768$ to $5248\times 3936$ , to validate the superiority of our method on a public benchmark. Of these, 400 images are for training, 50 and 150 images are used as validation and test data, respectively. Besides the above two datasets, we also employ the \textbf{BIG}~\cite{cascadePSP} dataset to evaluate the performance. The BIG dataset contains 150 images whose resolution ranges from $2048\times 1600$ to $5000\times 3600$, the annotations of image keeps the same guidelines as PASCAL VOC 2012~\cite{pascal}. Since the BIG dataset is very tiny, we use this dataset for testing only to evaluate the generality of the trained models.

Besides the high-resolution datasets, we also conduct experiments on widely used superpixel benchmarks, \textbf{BSDS500}~\cite{BSDS500} and \textbf{NYUv2}~\cite{NYUV2} to further validate the effectiveness of our proposed method. BSDS500 comprises 200 training, 100 validation and 200 test images, and each image is annotated by multiple semantic labels from different experts. To make a fair comparison, we follow previous works~\cite{SCN,SSN,SEAL} and treat each annotation as an individual sample. Consequently, 1,087 training, 546 validation samples and 1,063 testing samples could be obtained. NYUv2 is an indoor scene understanding dataset and contains 1,449 images with object instance labels. To evaluate the superpixel methods, Stutz \etal~\cite{SP_evaluation} remove the unlabelled regions near the boundary and collect a subset of 400 test images with size 608$\times$448 for superpixel evaluation. 

\begin{figure}[t]
\setlength{\abovecaptionskip}{-2pt} 
\begin{center}
        \includegraphics[width=3.0in]{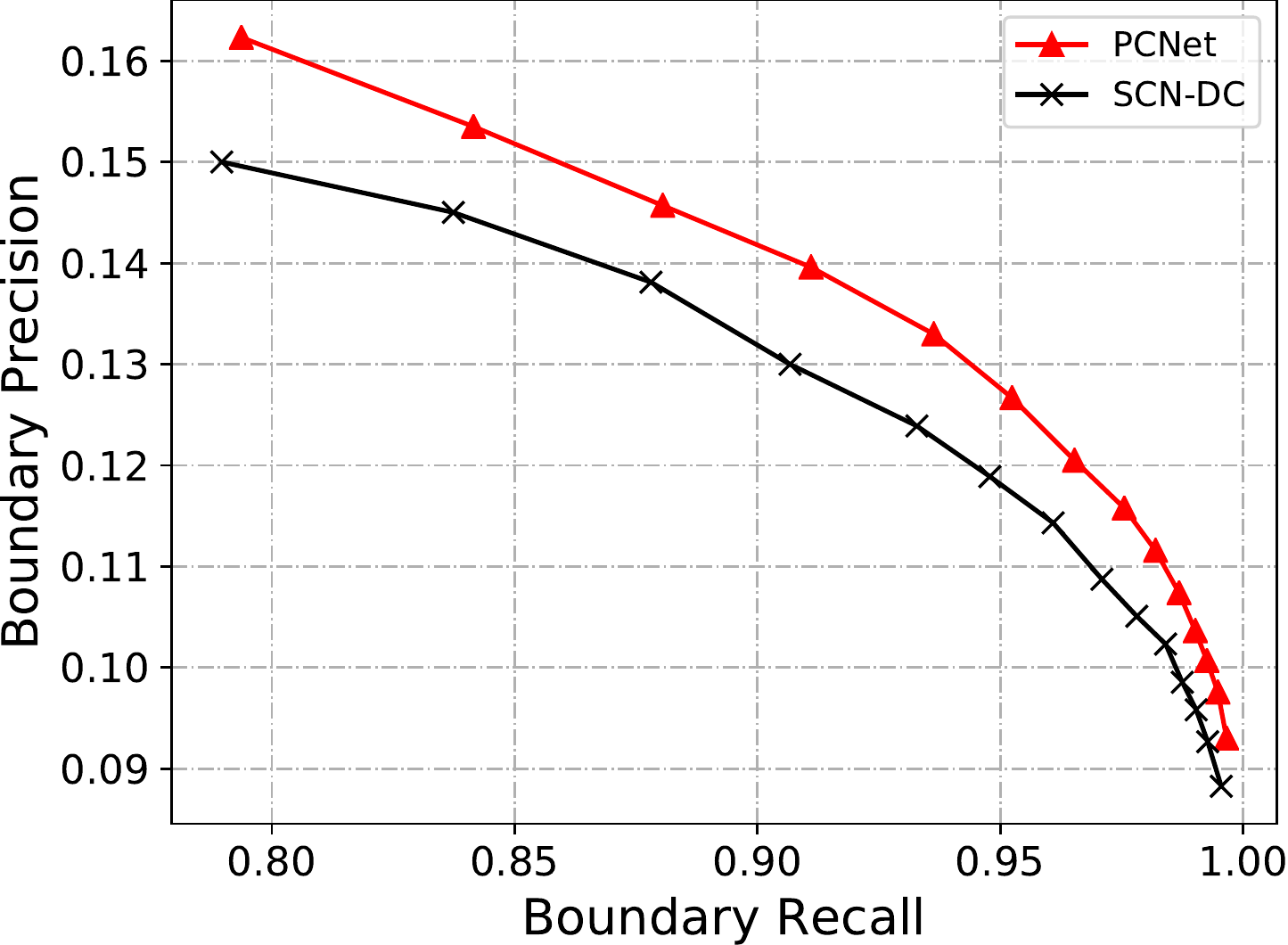}\\
\end{center}
\vspace{-0.15cm}
 \caption{Performance Comparison of our PCNet and the SCN~\cite{SCN} with divide-and-conquer strategy.}
 \vspace{0.3cm}
\label{dc_perform}
\end{figure}

\begin{figure}[t]
\setlength{\abovecaptionskip}{-2pt} 
\begin{center}
    \subfigure{
    \begin{minipage}[t]{0.8\linewidth}
        \includegraphics[width=2.44in]{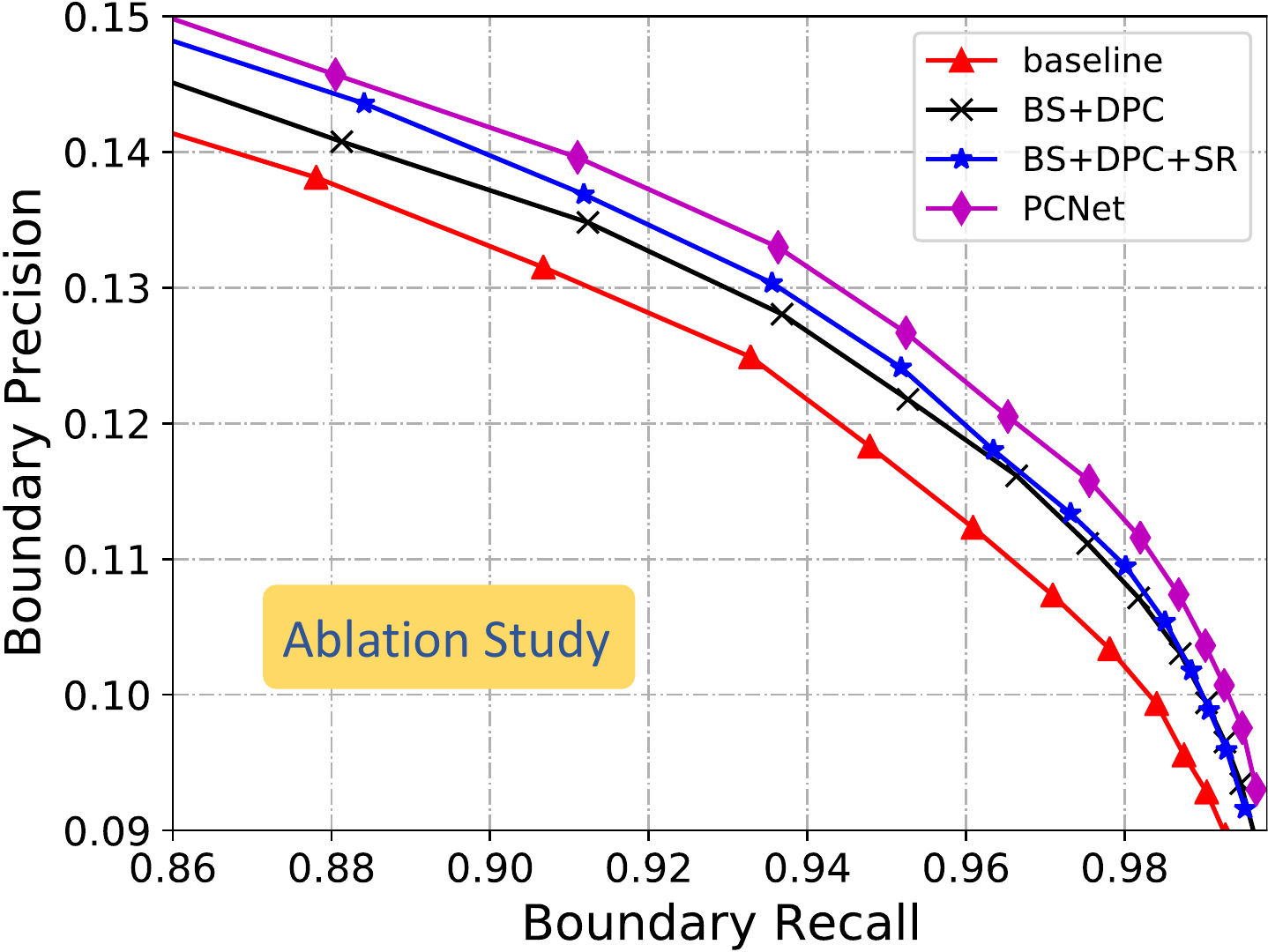}\\
    \end{minipage}
    }
    \\
    \vspace{-0.6cm}
    \subfigure{\begin{minipage}[t]{0.8\linewidth}
    \includegraphics[width=2.44in]{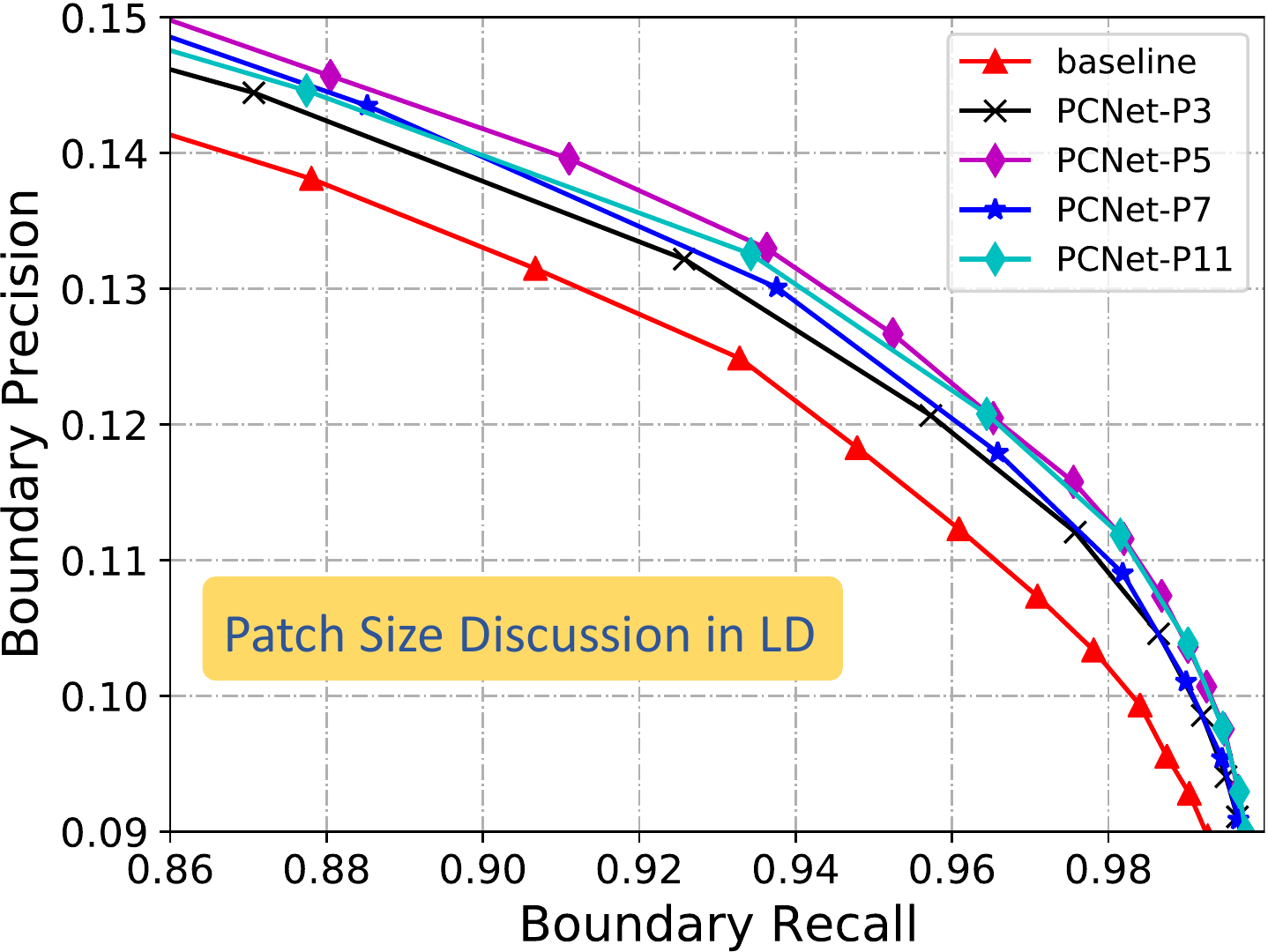}\\
    \end{minipage}}
\end{center}
\vspace{-0.3cm}
 \caption{Ablation study and the patch size discussion in local discrimination loss on Face-Human dataset, where the BS indicates the baseline method, SR stands for superpixel resolution branch, the PCNet-P\# means the PCNet equipped with  LD loss of patch size \#. }
 \vspace{0.6cm}
\label{abla}
\end{figure}

\subsection{Implementation Details} 
During training, 
we use the inputs with $128\times 128$  to predict the association maps and reconstructions for $512\times 512$ images. We first resize the original high-resolution image to $768\times 768$ and randomly crop a $512\times 512$ patch as the global samples $G^*$, while the local patch  $L^*$ obtained by directly cropping a $512\times 512$ patch from the original images. Both the global and local samples are down-sampled 4 times to serve as the global and the local inputs, respectively.  The encoder comprises 5 blocks, each block except the first down-samples the resolution by 2 times using the convolution with stride 2, while the decoder first restores the resolution as $128\times 128$ by the deconvolution operations, and a sub-pixel convolution is followed to output a $512\times 512$ feature map, which is further fed into the super-resolution and the superpixel prediction heads to reconstruct the image and give the association prediction. It is worth noting that the super-resolution branch is discarded during inference. And our local discrimination loss is performed on the pixel embedding $E$ in main branch to highlight the boundary pixels, we set the patch size as 5, \ie $K=5$. The hyperparameters, $\alpha,\beta$, are set as 0.1, and 0.5, respectively. The networks are trained using the adam optimization~\cite{adam} for 4k iterations with batch size 8, and start training with initial learning rate 5e-5, which is discounted by 10 every 2K iterations. When inference, for a test image, we first down-sample it by 4 times and feed it into the network to produce the association with the original resolution.

Several methods are considered for performance comparison, including classic methods, SLIC~\cite{SLIC}, LSC~\cite{LSC}, ERS~\cite{ERS}, SEEDS~\cite{SEEDS}, SNIC~\cite{SNIC} and the state-of-the-art deep method, SCN~\cite{SCN}. We simply use the OpenCV implementation for methods SLIC, LSC, and SEEDS. For other methods, we use the official implementations from the authors. As for another excellent method SSN~\cite{SSN}, since it could only process 1K resolution images in our practice,  therefore, we don't include this method for comparison.

\subsection{Comparison with the state-of-the-arts}
It is much important for superpixel segmentation to accurately identify the boundaries, therefore, we use the boundary recall (BR) and boundary precision (BP) to evaluate the performance~\cite{SP_evaluation}. 

To thoroughly evaluate the model, we train the network on Mapillary-Vistas or Face-Human datasets, and test on all three benchmarks to access the ability and generality of the models, especially for the deep model, SCN~\cite{SCN} and our proposed PCNet.
Fig.~\ref{main_perform} exhibits the BR-BP curves on three benchmarks, where the top row shows the performance comparison of all models trained on Face-Human and evaluated on three test sets, while the bottom row analogously exhibits the comparison on Mapillary-Vistas dataset.  Benefiting from the  differentiable convolution neural network, the deep learning methods, SCN and PCNet, perform much better than the traditional methods, SLIC~\cite{SLIC}, SNIC~\cite{SNIC}, SEEDS~\cite{SEEDS}, and ERS~\cite{ERS}.
Comparing with the state-of-the-art method, SCN, our PCNet performs slightly better on Face-Human and comparable on Mapillary-Vistas dataset. For the generality, our method perform better on the BIG dataset from Fig.~\ref{main_perform} I(c) and II(c). When generalizing to the Face-Human or Mapillary-Vistas, PCNet is comparable with SCN. 

Besides the high resolution datasets, we also conduct experiments on two widely-used datasets, BSDS500~\cite{BSDS500} and NYUv2~\cite{NYUV2} whose samples are with regular size. Following Yang~\cite{SCN} and Jampani~\cite{SSN}, we train the model on the BSDS500 dataset and test on on BSDS500 and NYUv2 dataset to evaluate the performance and the generality. The results are reported in Fig.~\ref{BDS_NYU}, we can see that the performance our PCNet on BSDS500 is still comparable with the model SCN, while performs worse than the SSN. When generalized to the NYUv2 dataset, our model could  achieve a slightly better performance that the SOTA methods SCN and SSN, which validates the effectiveness of our PCNet. What's more, our PCNet could also achieve a more outstanding inference efficiency due the lower resolution input.

Fig.~\ref{main_perform} suggests that our PCNet could achieve a slightly better or comparable performance with SCN using only a $1/4$ resolution version of the test image. The lower resolution input could not only allow the network to process very high-resolution \textbf{5K} images but significantly accelerate the inference due to the fewer float calculation during forwarding, just as shown in Fig.~\ref{fig1}.
Fig.~\ref{main_visual} subsequently visualizes the superpixel results on Face-Human, Mapillary-Vistas and Big datasets from top to bottom rows, intuitively showing the superiority of our method.

\subsection{Ablation Study}
In this subsection, we will first discuss about the division-and-conquer strategy to further clarify the motivation of our framework, then the contribution of each component in our PCNet is validated by a series of experiments.

\subsubsection{Discussion about the division-and-conquer strategy}
Since superpixel generation is an over-segmentation representation, we can generate superpixels by divide-and-conquer strategy for high-resolution images. For example, we can divide the input image as four non-overlapping sub-images, and run superpixel generation for each sub-image to get a superpixel label map. Then, we can combine the four maps by re-indexing each map to get the superpixels for the complete image. Intuitively, the division-and-conquer strategy is an straightforward choice for high-resolution superpixel segmentation. Actually, at the beginning of our practice, we have tried the division-and-conquer solution using the SOTA SCN model, but the results are not satisfactory. The first weakness of this strategy is that There would be an obvious line between patches; Sincethe  patch  boundary  blocks  the  superpixel  merging.   if  a compact  region  is  split  during  patch  division,  it  is  not possible  to  form  a  superpixel  for  the  pixels in  different patches, as shown in Fig.~\ref{dc_show}. Besides, the performance of division-and-conquer is also unsatisfactory. As shown in Fig.~\ref{dc_perform}, although the boundary recall is acceptable, the accuracy is too low.   Therefore, we abandon this  na\"\i ve strategy, and propose our PCNet to generate superpixels in an end-to-end fashion, which also achieves much better performance. 

\subsubsection{Validate the contribution of each component}
To validate the contribution of each module in PCNet, we conduct ablation study on Face-Human dataset to study their respective contribution including the DPC branch, super-resolution branch, and the local discrimination loss. 

The results are reported in Fig.~\ref{abla}, the baseline (BS) method is the na\"\i ve strategy shown in Fig.~\ref{argue}(b), which only employs the global image and uses the loss function Eq.~\ref{sp_loss} to train the network. As shown in Fig.~\ref{abla}, the baseline method performs very poorly, since the very low-resolution input sacrifices too many textures, especially the boundary contexts. With our DPC branch, more detailed boundaries could be perceived, consequently, the performance gets improved. The super-resolution branch further enables the network to restore more details from the low-resolution input. From Fig.~\ref{abla}, the performance steps further when the super-resolution branch is equipped. Fig.~\ref{binary_mask_validation} also suggests that our dynamic guiding training is more outstanding that the na\"\i ve choice, \ie, the ground-truth multi-class label, which validates the effectiveness of our dynamic guiding training mechanism.  When the local discrimination loss is further applied, we could achieve our best results.

Besides the validation of each component in our method, we also give a discussion about the performance effects of the patch size in our LD loss. In our standard PCNet, we sample $5\times 5$ patches in the baseline LD loss. In this section, we vary the patch size from 3 to 11 to study the performance difference.  The results are reported in the right figure of Fig.~\ref{abla}, where the PCNet-P\# indicates the PCNet with different patch sizes in LD loss. From Fig.~\ref{abla}, the LD loss with smaller patch size 3 contributes fewer performance gains. When the patch size increases to 5, the performance gets improved. However, further enlarging the patch size to 7 or 11  does not make the performance step further but perform a litter worse than PCNet-P5. We guess the reason stems from that the too large patch size would introduce more pixels that are not close enough to the boundary, leading to the LD loss fails to well focus on the boundary contexts.  Therefore, we set the patch size as 5 in our PCNet.

\section{Conclusion}
\label{conclusion}
This work proposes the first framework to perform the ultra high-resolution superpixel segmentation. For memory and computation efficiency, we employ a low-to-high prediction strategy that produces a high-resolution association map from a low-resolution input. Consequently, our proposed PCNet could tolerate higher resolution images and infer with a faster speed. To compensate boundary details lost in the down-sampled inputs, we design a decoupled patch calibration branch to calibrate the boundary pixel assignment of the global output.
A dynamic guiding mask is further proposed to enforce the DPC branch to focus on perceiving the boundaries.
To accurately identify more boundary pixels, we propose a local discrimination loss to highlight the pixel embeddings around the boundaries.
We conduct extensive experiments on two public benchmarks and our collected Face-Human dataset to evaluate the performance, our proposed PCNet could efficiently process very high-resolution 5K images while maintain the comparable performance with the state-of-the-art SCN.

%
\IEEEpeerreviewmaketitle

\nocite{*}


%

\end{document}